\documentclass[10pt]{article}

\usepackage[margin=1in]{geometry}
\usepackage{times}
\usepackage[numbers,sort&compress]{natbib}
\usepackage{authblk}
\usepackage{multirow}
\usepackage{needspace}
\usepackage[utf8]{inputenc}
\usepackage[T1]{fontenc}
\usepackage{hyperref}
\usepackage{url}
\usepackage{booktabs}
\usepackage{amsfonts}
\usepackage{nicefrac}
\usepackage{microtype}
\usepackage{xcolor}
\usepackage{graphicx}
\usepackage{wrapfig}
\usepackage{adjustbox}
\usepackage{amsthm}
\usepackage{subcaption}

\newtheorem{proposition}{Proposition}
\usepackage{algorithm}
\usepackage{amsmath,amssymb}
\usepackage{algorithmic}

\title{SPR: Toward a Graph Foundation Model for Transferable Graph Cognition via Spectral Patterns and Relational Geometry}

\author[1]{Ankang Yang$^{\dagger}$}
\author[1]{Jitao Zhao$^{\dagger}$}
\author[1]{Dongxiao He$^{*}$}
\author[1]{Di Jin}
\author[2]{Liang Yang}
\author[3]{Weixiong Zhang}

\affil[1]{School of Computer Science and Technology, Tianjin University, Tianjin, China\\
\texttt{\{yak123,zjtao,hedongxiao,jindi\}@tju.edu.cn}}
\affil[2]{School of Artificial Intelligence, Hebei University of Technology, Tianjin, China\\
\texttt{yangliang@vip.qq.com}}
\affil[3]{Department of Health Technology and Informatics; Department of Data Science and Artificial Intelligence,\\
The Hong Kong Polytechnic University, Kowloon, Hong Kong\\
\texttt{weixiong.zhang@polyu.edu.hk}}

\date{}

\begin{document}

\maketitle

\begingroup
\renewcommand{\thefootnote}{\fnsymbol{footnote}}
\footnotetext[2]{Equal contribution.}
\footnotetext[1]{Corresponding author.}
\endgroup

%(Graph foundation models aim to learn representations that generalize across diverse graphs 第一句话视野不开阔，而且和后面的不太匹配)（中间要有一个总结性的，这样后面的才是有逻辑的，比如目前已经有很多工作被提出，然而一个根本性问题，图中可迁移的机制尚未清晰）. （Yet existing approaches often characterize graphs at predetermined structural scales. Fixed-hop neighborhoods, predefined propagation schemes, and discrete structural units impose specific ways of describing a graph, making continuous multi-scale patterns difficult to model. Moreover, treating structural patterns as isolated units obscures their pairwise organization within each graph, making recurring relational organization across graphs difficult to discover.这一部分讲问题太快了，没过度，每一个大概形式要一样，总结的不够好） Effective cross-graph generalization therefore calls for modeling （both continuous multi-scale patterns and fine-grained relational geometry这两个要再提提炼一下，现在有点像一个细节，而不是一个总结性的东西） across graphs. We propose SPR, a graph foundation model that learns Spectral Patterns And Relational geometry. （Spectral parsing learns shared Chebyshev response functions on graph-specific normalized adjacency matrices, revealing continuous, multi-scale spectral patterns. In parallel, Gromov--Wasserstein prototype learning distills diffusion-induced pairwise geometry from multiple source graphs into a shared prototype geometry.这个要重点说我是怎么创新性的解决我上面提到的问题的，重点是创新性，而不是我具体怎么做的） This reveals recurring relational organization across graphs. SPR further refines the learned representations through local, diffusion, and prototype-guided interactions. Experiments demonstrate consistent impr

\begin{abstract}
Recently, Graph Foundation Models (GFMs) have attracted increasing attention for their potential to learn unified and generalizable knowledge across diverse graphs, thereby supporting a wide range of graph scenarios. However, unlike natural language and images, graphs lack an intuitive and unified form for perceiving and organizing transferable knowledge. Therefore, despite many initial explorations, a key problem remains unresolved: the transferable cognitive mechanism for graphs. Existing GFMs usually rely on intuitively defined mechanisms to encode transferable graph patterns, such as handcrafted structural templates (e.g., cycles and trees), fixed propagation mechanisms, and predefined random-walk patterns. These mechanisms are often prescriptive and rigid, limiting their ability to flexibly characterize diverse graph patterns across domains. This motivates the exploration of a more flexible, graph-native, and transferable cognitive mechanism. To this end, we analyze graphs from the spectral perspective and propose SPR. SPR decomposes graph information into Chebyshev polynomial bases and learns shared spectral responses, enabling unified yet adaptive cognition of continuously varying spectral patterns across graphs. We further model cross-graph relational geometry to distill recurring pairwise organization from multiple graphs, providing a shared relational reference for transferring such graph cognition across domains. Extensive experiments across diverse graphs and downstream scenarios demonstrate that SPR enables effective and transferable graph cognition.
\end{abstract}

\section{Introduction}

%Graph data → GNN → GFM → Existing limitation → Our observation on structural propagation → SPR

Graph data are prevalent in diverse real-world scenarios, such as recommender systems~\citep{DBLP:conf/kdd/YingHCEHL18} and social networks~\citep{DBLP:conf/kdd/PerozziAS14}, where entities interact through complex relations. Graphs provide a natural way to model such relational data by treating entities as nodes and their interactions as edges~\citep{Survey-GNNS,GNNsurvey2}. To learn from such data, Graph Neural Networks (GNNs) have become a dominant paradigm by leveraging message passing to aggregate information from neighboring nodes~\citep{DBLP:conf/iclr/KipfW17}. Yet conventional GNNs are typically developed for specific graphs or tasks, limiting their ability to reuse learned knowledge across different graph domains. Inspired by the transferability of foundation models in language and vision, recent research has moved toward Graph Foundation Models (GFMs), which aim to acquire transferable knowledge from multiple graphs and generalize it to diverse graph scenarios.~\citep{wang2025graph}.

%Following this trend, existing GFMs have explored several directions toward more generalizable graph learning. A major line of research focuses on large-scale graph pre-training to acquire knowledge that can be reused across graphs and tasks. Meanwhile, other studies seek to reduce the heterogeneity among different graphs, tasks, and feature spaces through various forms of  shared modeling. In addition, adaptation mechanisms, such as fine-tuning and graph prompting, have been developed to effectively transfer pre-trained knowledge to diverse downstream scenarios~\citep{GFMSurvey_Position,wang2025graph}. Together, these efforts have advanced the transferability of graph models beyond individual graphs and tasks.

Following this trend, recent GFMs have explored several  routes to enable knowledge sharing across diverse graphs. One direction seeks to unify graph semantics, typically by translating node and edge attributes into a common semantic space through textualization or language-model-based encoding, thereby reducing the incompatibility of feature spaces across domains~\citep{liu2023one,he2025unigraph}. A second line focuses on structural transfer, where graph-specific topology is abstracted into reusable structural units or encodings, such as computation-tree vocabularies, task trees, universal structural encodings, and other topology-derived descriptions~\citep{wang2024gft,DBLP:conf/icml/WangZMCZ025,DBLP:conf/www/ChenZWMLY26}. In parallel, another stream aims to bridge different graph tasks by reformulating node-, edge-, and graph-level prediction into shared learning interfaces and adapting pre-trained knowledge through graph prompting, instruction tuning, or lightweight task-specific modules~\citep{DBLP:conf/www/LiuY0023,liu2023one,he2025unigraph}. Despite their different formulations, these approaches share a common design principle: they first define a reusable form of graph knowledge and then build cross-graph pre-training and adaptation around it.

%Recent GFMs have increasingly moved beyond simply pre-training a shared encoder on multiple graphs toward explicitly modeling knowledge that can persist across graphs. The resulting challenge is to identify patterns that remain reusable across different topologies, domains, and tasks. Recent studies have  explored transferable graph knowledge in graph-native forms, including recurring structural patterns, invariant graph relations, shared geometric representations, and reusable structural behaviors~\citep{DBLP:conf/icml/WangZMCZ025,DBLP:journals/corr/abs-2603-00618,DBLP:journals/corr/abs-2605-08689,DBLP:journals/corr/abs-2606-03270,DBLP:journals/corr/abs-2601-22384}. Although these approaches differ substantially in formulation, they reflect a common shift in GFM research: （graph structure is no longer treated only as an input to be encoded, but as a source of knowledge that should be preserved and reused across learning contexts. How transferable graph patterns are represented therefore becomes critical to cross-graph generalization为什么要强调structure呢，这个可以不用专门提structure，就算落在structure上，也需要更仔细考虑说法，这个更像是已有的insight，不合适放在对相关工作的总结）.

Despite these initial explorations, a key problem remains unresolved: the transferable cognitive mechanism for graphs. Unlike natural language and images, graphs are less intuitive for humans to understand and lack an obvious common form through which their underlying patterns can be perceived and organized. This difficulty is further amplified by substantial variations in domain-specific semantics and graph topology~\citep{Survey-GNNS,GNNsurvey2}. Consequently, cross-graph transfer requires more than learning a shared representation; it calls for a graph-native principle that determines how graph knowledge should be perceived, identified and organized as transferable across different domains. We refer to such a principle as a transferable cognitive mechanism for graphs. Existing GFMs have instantiated such mechanisms through different intuitively defined priors, including predefined structural vocabularies composed of trees and cycles, fixed propagation mechanisms, and random-walk-based structural descriptions~\citep{DBLP:conf/www/0008HZW0Y25,DBLP:journals/corr/abs-2607-28980,DBLP:journals/corr/abs-2510-01510}. Despite their different forms, these mechanisms prescribe particular ways of perceiving and encoding graph knowledge, leaving the underlying cognitive mechanism largely determined by human-designed priors.

%（However, many existing GFMs still encode transferable structure through fixed neighborhoods, predefined propagation schemes, or discrete structural units. Such designs restrict structural modeling to a small set of prescribed scales,这个写的太小了，让人感觉不那么重要） limiting the ability to describe patterns that vary continuously across scales. The spectral domain offers a natural way to address this limitation, as spectral response functions characterize （graph signals 前文没提到过就不要用） along the frequency axis and provide a continuous multi-scale view of structural variation. Graph signals, however, can exhibit markedly different spectral characteristics~\citep{zhang2023spectral,DBLP:conf/iclr/DongSHWL25}. As shown in Figure~\ref{fig:raw_proxy_bands}, raw node features concentrate their energy in different frequency ranges across datasets, suggesting that a few fixed structural scales may be insufficient to characterize such variation. Beyond scale coverage, another challenge concerns the organization of structural patterns. Existing methods often identify reusable neighborhoods, motifs, or substructures as individual units but the relations among these units receive much less attention. The same set of local patterns can induce very different graph structures depending on their relative arrangement. Modeling this relational organization is therefore important for learning transferable structure.（motivation单独一段，而不是和你的解决思路混在一起，motivate对问题的描述不清晰，对解决的难题写到不充分，对方法的新颖性和巧妙也写的不好） 这个改一下

\begin{wrapfigure}{r}{0.5\textwidth}
  \vspace{-0.8em}
  \centering
  \includegraphics[width=0.48\textwidth]{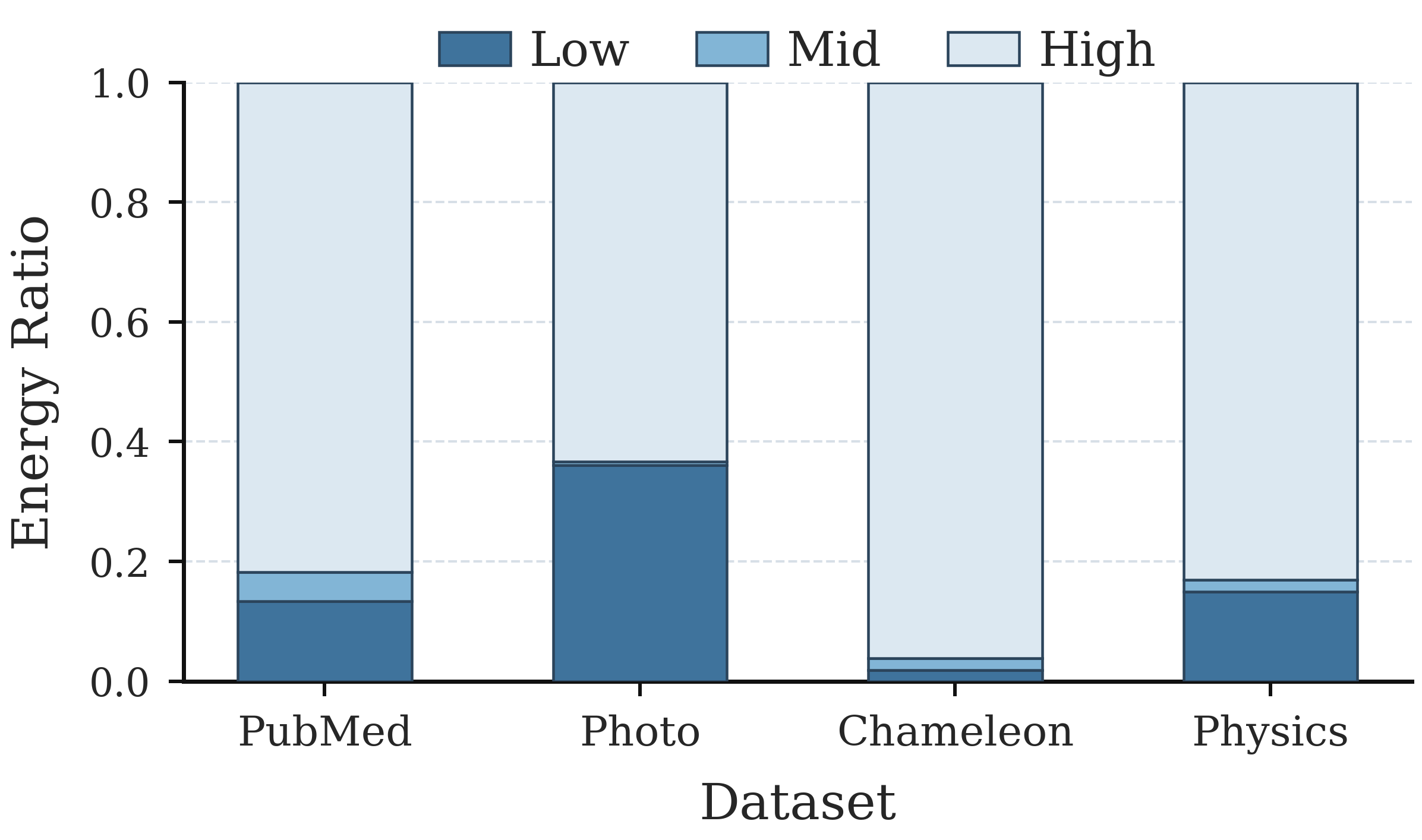}
  \caption{Spectral energy distribution of raw node features.}
  \label{fig:raw_proxy_bands}
  \vspace{-1em}
\end{wrapfigure}

This issue poses two key challenges: identifying an appropriate transferable cognitive mechanism for graphs, and enabling such a mechanism to be learned through multi-graph pre-training. The first challenge concerns what form this mechanism should take. Rather than being confined to a small set of predefined graph patterns, it should be sufficiently flexible to accommodate patterns that vary across different domains. As shown in Figure~\ref{fig:raw_proxy_bands}, raw node features exhibit markedly different spectral distributions across datasets~\citep{zhang2023spectral,DBLP:conf/iclr/DongSHWL25}, suggesting that informative patterns may occupy different and continuously varying spectral regions rather than a few predetermined resolutions. The second challenge concerns how transferability can emerge from pre-training rather than being manually imposed in advance. Learning from multiple graphs should therefore go beyond fitting each graph independently and uncover common organization across them. In particular, similar patterns may play different roles depending on their relations to other patterns, making relational organization an important basis for cross-graph transfer.

To address these challenges, we propose SPR, a graph foundation model that develops a flexible and transferable cognitive mechanism from the spectral perspective. For the first challenge, the spectral domain provides a continuous space in which graph patterns with different characteristics can be expressed under a unified view, avoiding the need to prescribe a small set of discrete patterns or propagation scales. Based on this perspective, SPR decomposes graph information into Chebyshev polynomial bases and learns shared spectral responses, enabling continuously varying spectral patterns to be adaptively perceived across graphs. For the second challenge, SPR further learns cross-graph relational geometry during multi-graph pre-training. Rather than modeling the organization within each graph independently, it characterizes intrinsic pairwise relations in individual graphs and distills their common relational organization into a shared prototype geometry. This geometry provides a common relational reference that can be instantiated on different graphs, thereby supporting the transfer of learned graph cognition to unseen domains.

Our contributions are summarized as follows:

\begin{itemize}

\item We investigate transferable cognitive mechanisms for graphs and introduce a spectral perspective for understanding graph knowledge across domains, providing a flexible and graph-native view of diverse graph patterns.

\item We instantiate this perspective in SPR through shared Chebyshev spectral responses that adaptively capture continuously varying graph patterns, together with cross-graph relational geometry that distills shared pairwise organization into a prototype geometry for knowledge transfer.

\item Extensive experiments on diverse source and unseen graphs, covering both node- and graph-level tasks, demonstrate the effectiveness and transferability of SPR.

\end{itemize}

\section{Related Work}

\paragraph{Graph Foundation Models.}
Graph foundation models (GFMs) aim to learn reusable knowledge that can generalize across graphs and tasks. Early studies focus on reducing discrepancies in feature spaces and task formulations through unified representations, language-based encoders, or shared pre-training frameworks~\citep{OFA,he2025unigraph,zhao2024all}. Subsequent methods place increasing emphasis on structural transfer, introducing reusable structural vocabularies, graph prompts, structural tokens, and adaptive architectures to capture patterns that persist across different graphs~\citep{wang2024gft,yu2025samgpt,chen2025autogfm}. More recent work further explores graph-native forms of transferable knowledge, including recurring structural patterns, topology-aware representations, and shared geometric spaces~\citep{sun2025riemanngfm,DBLP:journals/corr/abs-2603-00618,DBLP:journals/corr/abs-2605-08689}.

\paragraph{Spectral Graph Learning.}
Spectral graph learning represents graph signals through functions defined over the spectra of graph adjacency or Laplacian matrices~\citep{DBLP:journals/corr/BrunaZSL13,DBLP:journals/pieee/OrtegaFKMV18}. Early spectral GNNs parameterize filters directly in the spectral domain, while polynomial approximations such as Chebyshev filters avoid eigendecomposition and enable efficient localized computation~\citep{defferrard2016convolutional}. Later work develops adaptive polynomial filters and learnable frequency responses to better accommodate different graph structures and signal characteristics~\citep{zhang2023spectral,DBLP:conf/iclr/DongSHWL25}. Recent studies further explore multi-resolution spectral bases and more flexible spectral parameterizations; for example, WaveGC learns general graph spectral wavelets through Chebyshev-order decomposition to model information over different spatial ranges~\citep{DBLP:conf/icml/0001H0GBB25}.

\paragraph{Gromov--Wasserstein for Graph Learning.}
Gromov--Wasserstein (GW) distance compares relational spaces through their intrinsic pairwise structure without requiring explicit correspondence between their elements~\citep{memoli2011gromov,peyre2016gromov}. Early graph applications mainly use GW for graph matching, comparison, and soft structural alignment~\citep{titouan2019optimal}. Its use has since expanded to graph coarsening, representation alignment, and structure-aware learning across graphs with different node sets or feature spaces. Recent work further applies GW to cross-graph transfer: GTRANS aligns and transfers structural patterns between source and target graphs through GW optimal transport, while related methods use fused GW to align graph representation spaces in distributed learning settings~\citep{DBLP:conf/nips/WangCMC25,DBLP:conf/nips/FuLHDZPC25}.

\section{Preliminaries}

\paragraph{Gromov--Wasserstein Geometry.}
Gromov--Wasserstein (GW) geometry compares relational spaces through their
intrinsic pairwise structure without requiring element-wise correspondence
~\citep{memoli2011gromov}.
For two graphs represented by structural cost matrices
$C_1\in\mathbb{R}^{n_1\times n_1}$ and
$C_2\in\mathbb{R}^{n_2\times n_2}$, with node distributions
$p\in\Delta^{n_1}$ and $q\in\Delta^{n_2}$, the discrete GW discrepancy is
\begin{equation}
\mathrm{GW}(C_1,C_2,p,q)
=
\min_{T\in\Pi(p,q)}
\sum_{i,i',j,j'}
\left|C_1(i,i')-C_2(j,j')\right|^2
T_{ij}T_{i'j'},
\label{eq:prelim_discrete_gw}
\end{equation}
where
$\Pi(p,q)=\{T\geq0\mid T\mathbf{1}=p,\,
T^\top\mathbf{1}=q\}$.
The transport $T$ provides a soft correspondence between the two spaces, while
the objective favors correspondences that preserve pairwise relations.
This property makes GW suitable for comparing graphs with different node sets
and feature spaces~\citep{DBLP:journals/focm/Memoli11,titouan2019optimal}.

\paragraph{Spectral Graph Filtering.}
Let $L=I-D^{-1/2}AD^{-1/2}=U\Lambda U^\top$ be the normalized graph Laplacian, where $\Lambda=\operatorname{diag}(\lambda_1,\ldots,\lambda_n)$ contains Laplacian frequencies. A spectral filter acts on a graph signal $X$ as
\begin{equation}
g_\theta(L)X
=
U g_\theta(\Lambda)U^\top X,
\label{eq:spectral_convolution}
\end{equation}
where $g_\theta$ specifies the response applied to different spectral
components~\citep{DBLP:journals/corr/BrunaZSL13,DBLP:journals/pieee/OrtegaFKMV18}.
Polynomial parameterizations such as Chebyshev filters approximate this response
without explicitly computing the eigendecomposition~\citep{defferrard2016convolutional}.
In SPR, the spectral frontend is implemented on the normalized propagation operator
$S=D^{-1/2}AD^{-1/2}=I-L$; hence a Laplacian frequency $\lambda$ corresponds to the
propagation-domain eigenvalue $\rho=1-\lambda$.

\section{Method}
\begin{figure*}[!t]
  \centering
  \includegraphics[
    width=\textwidth
  ]{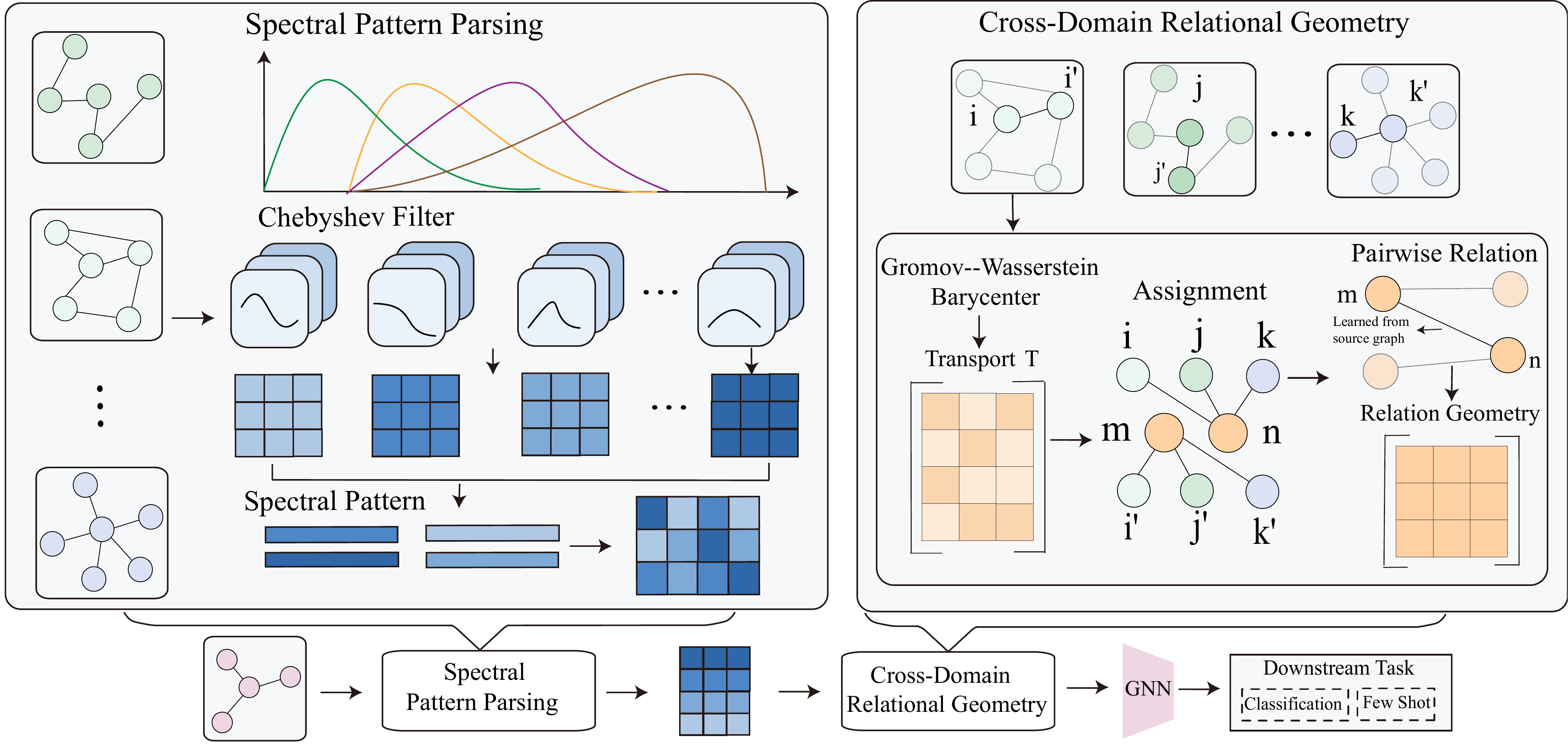}
  \caption{Overview of SPR.}
  \label{fig:overview}
\end{figure*}

SPR adopts a spectral view of transferable graph cognition with two components: spectral pattern parsing and cross-graph relational geometry. The former learns shared spectral responses instantiated on graph-specific operators, providing a unified yet adaptive representation across spectral regions. The latter aggregates intrinsic pairwise relations from source graphs into a shared prototype geometry. Both are integrated into a shared encoder for multi-graph self-supervised pre-training, as illustrated in Figure~\ref{fig:overview}.

\subsection{Spectral Pattern Parsing}

Spectral pattern parsing represents graph information over a continuous spectral axis rather than through fixed structural units or propagation scales. As Figure~\ref{fig:raw_proxy_bands} shows, different graphs exhibit distinct spectral characteristics. SPR therefore learns shared responses whose emphasis varies continuously over the spectrum, yielding a unified yet adaptive description across domains.

For a graph $G_m=(V_m,E_m)$ with node features $X_m\in\mathbb{R}^{n_m\times d_m}$, let $A_m$ and $D_m$ denote its adjacency and degree matrices. We use the normalized adjacency $S_m=D_m^{-1/2}A_mD_m^{-1/2}$ to characterize graph topology in the spectral domain. Instead of explicitly computing its eigendecomposition, SPR parameterizes spectral responses with Chebyshev polynomials, which can be evaluated efficiently through recursive propagation over $S_m$. Specifically, we construct the Chebyshev basis responses as
\begin{equation}
B_{m,0}=X_m,\qquad
B_{m,1}=S_mX_m,\qquad
B_{m,k}=2S_mB_{m,k-1}-B_{m,k-2},\quad k\geq2.
\label{eq:cheb_basis}
\end{equation}
Here, $B_{m,k}\in\mathbb{R}^{n_m\times d_m}$ denotes the $k$-th Chebyshev basis response and $K$ is the maximum polynomial order. Each basis corresponds to a different polynomial transformation of $S_m$. Rather than associating individual basis orders with predetermined structural scales, SPR learns how these responses should be combined.
For the $\alpha$-th spectral filter, we compute
\begin{equation}
Z_{m,\alpha}
=
\sum_{k=0}^{K}
\theta_{\alpha,k}B_{m,k},
\qquad
\alpha=1,\ldots,M_f,
\label{eq:spectral_filter_response}
\end{equation}
where $\theta_{\alpha,k}$ denotes the learnable coefficient of the $k$-th basis, $M_f$ is the number of spectral filters, and $Z_{m,\alpha}$ is the response produced by the $\alpha$-th filter. The coefficients are shared across graphs.

The spectral interpretation of Eq.~\eqref{eq:spectral_filter_response} becomes clearer by considering
$S_m=U_m\operatorname{diag}(\rho_m)U_m^\top$, where $\rho_m$ denotes the propagation-domain eigenvalues. Although this eigendecomposition is not explicitly computed, the filtering operation is equivalently written as
\begin{equation}
Z_{m,\alpha}=U_m g_\alpha(\rho_m)U_m^\top X_m,\qquad
g_\alpha(\rho)=\sum_{k=0}^{K}\theta_{\alpha,k}T_k(\rho).
\label{eq:spectral_response_function}
\end{equation}

Here, $U_m$ contains the eigenvectors of $S_m$, $\rho$ denotes an eigenvalue of $S_m$, and $T_k(\rho)$ is the $k$-th Chebyshev polynomial. Since $S_m=I-L_m$, a Laplacian frequency $\lambda$ corresponds to $\rho=1-\lambda$. The function $g_\alpha(\rho)$ therefore specifies the response of the $\alpha$-th filter to different spectral components under the propagation-domain parameterization.
Because $g_\alpha$ varies continuously with $\rho$, multiple shared response functions can emphasize complementary spectral regions without manually specifying frequency bands. Their realizations remain graph-dependent through $S_m$ and $X_m$, allowing the same parameterization to transfer across graphs with different topology and feature spaces.

Since the filtered responses $Z_{m,\alpha}$ retain graph-specific feature dimensions, we apply the same truncated-SVD compression to each response and denote the resulting node-side rank-$r$ embedding by
$R_{m,\alpha}=\mathrm{SVD}^{\mathrm{node}}_r(Z_{m,\alpha})\in\mathbb{R}^{n_m\times r}$.
Here, $\mathrm{SVD}^{\mathrm{node}}_r(\cdot)$ denotes the $n_m\times r$ node representation returned by the rank-$r$ truncated decomposition, using the same convention for every graph. This step only standardizes dimensionality across domains; the transferable spectral patterns are defined by the shared responses before compression.
We then concatenate the reduced responses and project them into the encoder space:
\begin{equation}
X_{\mathrm{spec},m}
=
\operatorname{LN}
\left(
[R_{m,1}\Vert\cdots\Vert R_{m,M_f}]
W_{\mathrm{spec}}+b_{\mathrm{spec}}
\right).
\label{eq:parsed_spectral_representation}
\end{equation}

Here, $\Vert$ denotes concatenation, $W_{\mathrm{spec}}$ and $b_{\mathrm{spec}}$ are learnable projection parameters, and $\operatorname{LN}(\cdot)$ denotes layer normalization. The resulting representation integrates multiple learned spectral responses in a common feature space, providing a continuous and data-adaptive characterization of graph patterns across spectral regions.
\subsection{Cross-Graph Relational Geometry}

SPR further distills reusable relational organization from multiple source graphs through intrinsic pairwise geometry. Each graph is characterized by node-pair relations, whose common structure is aggregated into a shared prototype geometry. This avoids requiring aligned node identities or feature spaces and provides a common relational reference for transfer.

For each source graph $G_m$, we construct the heat kernel $H_m=\exp[-t_{\mathrm{gw}}(I-S_m)]$, where $t_{\mathrm{gw}}$ controls the diffusion scale. The corresponding diffusion-induced distance is
\begin{equation}
C_m(i,j)
=
\sqrt{
H_m(i,i)+H_m(j,j)-2H_m(i,j)
}.
\label{eq:heat_dist}
\end{equation}
Here, $C_m\in\mathbb{R}^{n_m\times n_m}$ collects pairwise diffusion distances, providing an intrinsic graph description that does not require aligned node identities or feature dimensions.

Given the source geometries $\{C_m\}_{m=1}^{M}$, we learn a shared prototype geometry
$C_\star\in\mathbb{R}^{p\times p}$ through a Gromov--Wasserstein barycenter:
\begin{equation}
C_\star
=
\arg\min_{C,\{T_m\}}
\sum_{m=1}^{M}\lambda_m
\sum_{i,i',k,l}
\bigl(C_m(i,i')-C(k,l)\bigr)^2
T_m(i,k)T_m(i',l),
\label{eq:gw_barycenter}
\end{equation}
where $p$ is the number of prototypes, $w_m$ and $b$ denote the node and prototype distributions, respectively, and
$T_m\in\Pi(w_m,b)\subset\mathbb{R}^{n_m\times p}$
is the corresponding soft transport. The coefficient $\lambda_m$ denotes the source-graph weight with
$\sum_m\lambda_m=1$~\citep{peyre2016gromov}. In practice, we optimize an entropically regularized GW approximation using the solver settings reported in Appendix~\ref{app:impl}.

GW aligns pairwise relations rather than individual nodes: if $(i,i')$ is softly associated with prototype pair $(k,l)$, the objective encourages $C_m(i,i')$ and $C_\star(k,l)$ to agree. Thus, $C_\star$ represents recurring relational organization across source graphs.

This relational transfer admits a direct distortion guarantee.

\begin{proposition}[Distortion-controlled geometry transfer]
\label{prop:gw_lifting}
Let $R_m=\mathrm{diag}(w_m)^{-1}T_m$ and
$\overline C_m=R_mC_\star R_m^\top$. Then
\begin{equation}
\|C_m-\overline C_m\|_{w_m\otimes w_m}^{2}
\leq
\mathcal{D}_{\mathrm{GW}}(C_m,C_\star,T_m).
\end{equation}
\end{proposition}

Thus, a small GW distortion guarantees that the shared prototype geometry can be
lifted back to each graph with bounded relational distortion.
The proof is provided in Appendix~\ref{app:proof_gw_lifting}.

To incorporate the learned geometry into graph representations, we convert the prototype distances into relational affinities.
Following our implementation, we define
\begin{equation}
A_\star(k,l)
=
\exp\left(
-C_\star(k,l) / s_\star
\right),
\qquad
s_\star=\operatorname{median}(C_\star),
\end{equation}
followed by symmetrization and removal of self-connections.
A diffusion kernel is then constructed over the resulting prototype graph:
\begin{equation}
K_\star
=
\exp\bigl(-L(A_\star)\bigr).
\label{eq:prototype_kernel}
\end{equation}
where $L(A_\star)$ is the graph Laplacian of the prototype affinity graph. The resulting $K_\star$ serves as the shared prototype interaction kernel.

For a source graph $G_m$, its correspondence with the shared prototype space allows the learned relational structure to be instantiated in the original node space:
\begin{equation}
\widetilde P_m
=
T_mK_\star T_m^\top .
\label{eq:prototype_projection}
\end{equation}
The composition transfers prototype interactions back to the graph node space; a symmetrically normalized form of $\widetilde P_m$ is used by the encoder.

The same shared geometry naturally extends to unseen graphs.
For a new graph $G_{\mathrm{new}}$, we retain $C_\star$ and $K_\star$ and estimate a graph-to-prototype assignment $T_{\mathrm{new}}$. We use GW alignment when the pairwise geometry is tractable; for large graphs, we use the scalable K-means-based sparse assignment described in Appendix~\ref{app:impl}. The transferred relational structure is
\begin{equation}
\widetilde P_{\mathrm{new}}
=
T_{\mathrm{new}}K_\star T_{\mathrm{new}}^\top .
\label{eq:new_graph_prototype_projection}
\end{equation}
The shared prototype geometry remains fixed during transfer. The distortion guarantee above applies to GW-based alignment; the large-graph assignment is an efficiency-oriented approximation and is not covered by that guarantee.

The encoder combines local adjacency, graph-specific heat diffusion, and the transferred prototype operator through a node-wise gate.
\subsection{Self-Supervised Pre-training}
\label{sec:self_supervised_pretraining}

SPR is pretrained across the source graphs with node-level contrastive learning and masked spectral reconstruction. The shared spectral responses and encoder parameters are optimized jointly across graphs, while the prototype geometry is distilled from their graph-specific relational structures. Given two augmented graph views, the contrastive objective encourages corresponding nodes to maintain consistent representations:
\begin{equation}
\mathcal{L}_{\mathrm{node}}
=
-\frac{1}{N}\sum_i
\log
\frac{
\exp(\mathrm{sim}(q_i,q_i')/\tau_c)
}{
\sum_j \exp(\mathrm{sim}(q_i,q_j')/\tau_c)
}.
\label{eq:node_contrastive_loss}
\end{equation}
where $q_i$ and $q_i'$ denote the representations of node $i$ from the two augmented views, $\mathrm{sim}(\cdot,\cdot)$ is the similarity function, and $\tau_c$ is the contrastive temperature.

To preserve the multi-scale information produced by spectral parsing, we hide the parsed spectral inputs of a node subset $\mathcal{M}$ before encoding and reconstruct their original, unmasked targets:
\begin{equation}
\mathcal{L}_{\mathrm{spec}}
=
\frac{1}{|\mathcal{M}|}
\sum_{i\in\mathcal{M}}
\left\|
f_{\mathrm{rec}}(z_i)-x_{\mathrm{spec},i}
\right\|_2^2,
\label{eq:spectral_reconstruction_loss}
\end{equation}
where $z_i$ is produced from the masked input, $x_{\mathrm{spec},i}$ is the corresponding unmasked spectral target, and $f_{\mathrm{rec}}$ is the reconstruction head.
The overall pre-training objective is
\begin{equation}
\mathcal{L}
=
\lambda_{\mathrm{node}}\mathcal{L}_{\mathrm{node}}
+
\lambda_{\mathrm{spec}}\mathcal{L}_{\mathrm{spec}}.
\end{equation}
where $\lambda_{\mathrm{node}}$ and $\lambda_{\mathrm{spec}}$ control the two objectives. The contrastive objective promotes representation consistency across graph perturbations, while spectral reconstruction encourages the encoder to preserve the spectral patterns extracted by spectral parsing.
Implementation details and loss weights are provided in Appendix~\ref{app:impl}.

\section{Experiments}

We evaluate SPR on node- and graph-level classification across diverse graph domains, with particular emphasis on whether the learned graph cognition remains reusable beyond the pre-training graphs. For node classification, SPR is pretrained on Cora, Photo, CS, and Chameleon, and evaluated on unseen graphs including Computers, Physics, Citeseer, Pubmed, Cornell, Wisconsin, ogbn-arxiv, and ogbn-products~\citep{sen2008collective,DBLP:conf/iclr/PeiWCLY20,DBLP:journals/corr/abs-1811-05868,hu2020open}. For graph classification, we evaluate on COLLAB, DD, ENZYMES, IMDB-B, and PROTEINS~\citep{DBLP:journals/corr/abs-2007-08663}. We consider both a $1{:}1{:}8$ train/validation/test split and the 1-shot setting, providing complementary evaluations of self-supervised representation quality and label-efficient transfer. We compare SPR with self-supervised graph learning methods, including GRACE, BGRL, and GraphMAE2~\citep{zhu2020deep,thakoor2021bootstrapped,hou2023graphmae2}, as well as graph foundation models, including GCOPE, GraphAny, SAMGPT, TIG, and GraphGlue~\citep{zhao2024all,zhao2024fully,yu2025samgpt,DBLP:conf/www/ZhaoWLHJFZ26,DBLP:journals/corr/abs-2603-00618}. Additional implementation and evaluation details are provided in Appendix~\ref{app:impl}.

\subsection{1-shot Node and Graph Classification}

Table~\ref{tab:oneshot_all_results} reports the 1-shot results, where only one labeled example per class is available for downstream adaptation. This setting provides a stringent test of transferability because the limited supervision leaves little opportunity to relearn graph-specific patterns on the target domain.

\begin{table*}[!t]
\centering
\caption{1-shot performance comparison on node and graph classification. Results are accuracy (\%, mean $\pm$ std); best and runner-up results are bolded and underlined, respectively.}
\label{tab:oneshot_all_results}
\small
\setlength{\tabcolsep}{2.0pt}
\renewcommand{\arraystretch}{1.08}
\resizebox{\textwidth}{!}{
\begin{tabular}{lccc@{\hspace{6pt}}cccccc}
\toprule
& \multicolumn{3}{c}{\textbf{Self-supervised Pre-training}}
& \multicolumn{6}{c}{\textbf{Graph Foundation Models}} \\
\cmidrule(lr){2-4} \cmidrule(lr){5-10}
& BGRL & GraphMAE2 & GRACE
& GCOPE & GraphAny & TIG & SAMGPT & GraphGlue & \textbf{SPR} \\
\midrule

\multicolumn{10}{l}{\textbf{Source-Graph Node Classification}} \\
Cora
& $36.85 \pm 5.98$ & $44.14 \pm 9.22$ & $41.34 \pm 5.74$
& $\underline{50.59 \pm 9.05}$ & $39.40 \pm 8.41$ & $46.06 \pm 10.78$ & $39.65 \pm 7.27$ & $32.83 \pm 6.24$
& $\mathbf{53.29 \pm 10.12}$ \\
Photo
& $42.59 \pm 9.21$ & $57.88 \pm 9.49$ & $55.70 \pm 8.40$
& $49.90 \pm 7.57$ & $\underline{61.53 \pm 9.87}$ & $53.95 \pm 9.47$ & $55.32 \pm 9.01$ & $48.33 \pm 2.69$
& $\mathbf{64.63 \pm 10.19}$ \\
CS
& $41.05 \pm 7.42$ & $65.56 \pm 6.98$ & $60.72 \pm 6.94$
& $63.04 \pm 8.11$ & $70.04 \pm 6.99$ & $\underline{70.89 \pm 7.14}$ & $64.29 \pm 8.41$ & $67.43 \pm 9.08$
& $\mathbf{71.99 \pm 6.87}$ \\
Chameleon
& $21.16 \pm 2.72$ & $25.15 \pm 4.77$ & $23.52 \pm 3.34$
& $\underline{27.66 \pm 4.41}$ & $24.91 \pm 4.20$ & $24.96 \pm 4.82$ & $24.94 \pm 4.21$ & $21.68 \pm 1.79$
& $\mathbf{28.01 \pm 4.42}$ \\[0.4mm]

\specialrule{0.35pt}{1pt}{1pt}
\multicolumn{10}{l}{\textbf{Held-out Graph Node Classification}} \\
Citeseer
& $25.97 \pm 3.21$ & $38.86 \pm 8.85$ & $23.03 \pm 2.74$
& $39.58 \pm 8.29$ & $35.58 \pm 6.99$ & $\underline{40.55 \pm 8.78}$ & $34.58 \pm 7.12$ & $28.70 \pm 4.70$
& $\mathbf{45.44 \pm 9.68}$ \\
Pubmed
& $37.69 \pm 7.40$ & $51.56 \pm 8.81$ & $39.80 \pm 6.64$
& $48.32 \pm 8.36$ & $\underline{51.84 \pm 9.42}$ & $51.01 \pm 6.54$ & $49.20 \pm 8.52$ & $48.65 \pm 6.27$
& $\mathbf{55.73 \pm 11.65}$ \\
Physics
& $54.56 \pm 10.27$ & $\underline{78.76 \pm 9.14}$ & $50.44 \pm 10.98$
& $58.03 \pm 9.67$ & $75.01 \pm 11.41$ & $68.14 \pm 15.76$ & $61.55 \pm 13.04$ & $69.99 \pm 8.69$
& $\mathbf{83.84 \pm 7.66}$ \\
Cornell
& $26.47 \pm 9.91$ & $31.36 \pm 10.60$ & $26.47 \pm 7.49$
& $35.25 \pm 10.94$ & $36.22 \pm 9.52$ & $29.36 \pm 11.34$ & $30.46 \pm 7.77$ & $\underline{36.89 \pm 9.49}$
& $\mathbf{44.52 \pm 12.37}$ \\
Wisconsin
& $22.36 \pm 9.18$ & $33.61 \pm 10.91$ & $23.50 \pm 9.89$
& $\underline{37.25 \pm 11.32}$ & $23.32 \pm 8.61$ & $31.01 \pm 8.48$ & $32.19 \pm 7.27$ & $36.17 \pm 9.23$
& $\mathbf{52.81 \pm 15.06}$ \\
Arxiv
& $4.15 \pm 0.82$ & $14.46 \pm 3.47$ & $7.72 \pm 1.89$
& $7.47 \pm 1.43$ & $12.64 \pm 2.15$ & $2.62 \pm 1.01$ & $\underline{17.25 \pm 3.94}$ & $4.73 \pm 0.85$
& $\mathbf{23.04 \pm 6.06}$ \\
Products
& $11.21 \pm 2.39$ & $14.17 \pm 4.34$ & $13.07 \pm 2.21$
& $3.80 \pm 0.78$ & $10.97 \pm 4.48$ & $4.02 \pm 1.37$ & $\underline{15.16 \pm 2.72}$ & $3.71 \pm 0.86$
& $\mathbf{17.96 \pm 5.10}$ \\[0.4mm]

\specialrule{0.35pt}{1pt}{1pt}
\multicolumn{10}{l}{\textbf{Graph-Level Transfer}} \\
COLLAB
& $43.42 \pm 11.74$ & $42.64 \pm 7.97$ & $42.72 \pm 8.08$
& $38.06 \pm 6.97$ & -- & $\underline{50.27 \pm 1.19}$ & $46.46 \pm 10.57$ & $38.22 \pm 5.76$
& $\mathbf{53.68 \pm 10.60}$ \\
DD
& $51.14 \pm 4.23$ & $51.95 \pm 5.68$ & $51.69 \pm 4.22$
& $\underline{59.61 \pm 13.10}$ & -- & $52.08 \pm 0.34$ & $52.15 \pm 5.20$ & $51.81 \pm 5.32$
& $\mathbf{74.84 \pm 11.98}$ \\
ENZYMES
& $19.19 \pm 2.93$ & $20.07 \pm 3.72$ & $20.11 \pm 3.52$
& $19.75 \pm 3.66$ & -- & $20.02 \pm 0.54$ & $\underline{20.12 \pm 3.73}$ & $19.75 \pm 3.44$
& $\mathbf{35.66 \pm 5.28}$ \\
IMDB-B
& $53.44 \pm 6.67$ & $52.81 \pm 6.82$ & $53.09 \pm 6.23$
& $51.91 \pm 6.50$ & -- & $\underline{55.74 \pm 0.55}$ & $53.30 \pm 6.56$ & $53.56 \pm 6.04$
& $\mathbf{64.50 \pm 11.89}$ \\
PROTEINS
& $51.62 \pm 5.96$ & $53.37 \pm 9.13$ & $53.62 \pm 8.83$
& $\underline{54.52 \pm 8.71}$ & -- & $53.31 \pm 0.31$ & $53.41 \pm 9.29$ & $53.36 \pm 9.41$
& $\mathbf{65.46 \pm 7.14}$ \\

\bottomrule
\end{tabular}
}
\end{table*}

For node classification, SPR achieves the best mean performance on all source and unseen graphs. Its average margin over the runner-up increases from $1.81$ percentage points on source graphs to $6.52$ on unseen graphs, with clear gains on Wisconsin, Cornell, and Arxiv. This is consistent with SPR's shared-yet-adaptive design: spectral responses are shared but instantiated through graph-specific operators, while the prototype geometry is aligned to each target graph, remaining effective under minimal supervision.

The graph-classification results provide a further test of whether the learned knowledge remains reusable beyond node-level prediction. SPR achieves the best performance on all five benchmarks, outperforming the runner-up by $10.78$ percentage points on average. Since these datasets differ from the pre-training graphs and also require graph-level rather than node-level prediction, the consistent gains indicate that the learned representations are not tied to a particular source graph or prediction granularity. The results are also consistent with the role of relational geometry, which organizes pairwise relations in a shared prototype space that can be instantiated on new graphs.

\subsection{Self-Supervised Representation Learning}

Table~\ref{tab:official_all_results} evaluates representations learned through self-supervised pre-training using a frozen encoder and linear probing. SPR is pretrained on four source graphs and evaluated on source and unseen node-classification graphs, together with graph-level benchmarks. Because the pretrained encoder is fixed during downstream evaluation, this setting directly measures the quality and reusability of the learned representations without relearning the encoder for each dataset.

\begin{table*}[!t]
\centering
\caption{Self-supervised representation learning performance on node and graph classification. Results are accuracy (\%, mean $\pm$ std); best and runner-up results are bolded and underlined, respectively.}
\label{tab:official_all_results}
\small
\setlength{\tabcolsep}{2.0pt}
\renewcommand{\arraystretch}{1.0}
\resizebox{\textwidth}{!}{
\begin{tabular}{lccc@{\hspace{6pt}}cccccc}
\toprule
& \multicolumn{3}{c}{\textbf{Self-supervised Pre-training}}
& \multicolumn{6}{c}{\textbf{Graph Foundation Models}} \\
\cmidrule(lr){2-4} \cmidrule(lr){5-10}
& BGRL & GraphMAE2 & GRACE
& GCOPE & GraphAny & TIG & SAMGPT & GraphGlue & \textbf{SPR} \\
\midrule

\multicolumn{10}{l}{\textbf{Source-Graph Node Classification}} \\
Cora
& $81.22 \pm 1.17$ & $\underline{82.38 \pm 1.10}$ & $77.28 \pm 0.34$
& $79.97 \pm 0.50$ & $78.83 \pm 2.11$ & $81.19 \pm 1.26$ & $81.79 \pm 0.71$ & $77.54 \pm 0.64$
& $\mathbf{84.10 \pm 0.54}$ \\
Photo
& $89.48 \pm 0.93$ & $90.24 \pm 0.15$ & $91.51 \pm 0.23$
& $80.65 \pm 0.54$ & $87.48 \pm 3.44$ & $\underline{92.88 \pm 0.35}$ & $89.92 \pm 0.65$ & $87.57 \pm 0.40$
& $\mathbf{93.60 \pm 0.19}$ \\
CS
& $88.55 \pm 0.97$ & $91.82 \pm 0.09$ & $91.97 \pm 0.08$
& $86.13 \pm 0.20$ & $90.66 \pm 0.50$ & $\underline{92.62 \pm 0.26}$ & $88.76 \pm 0.20$ & $86.14 \pm 0.27$
& $\mathbf{93.45 \pm 0.26}$ \\
Chameleon
& $26.95 \pm 1.28$ & $44.18 \pm 2.90$ & $35.54 \pm 0.52$
& $\underline{53.10 \pm 0.83}$ & $\mathbf{57.50 \pm 1.47}$ & $34.52 \pm 2.80$ & $30.19 \pm 2.53$ & $24.26 \pm 3.48$
& $46.18 \pm 1.07$ \\

\specialrule{0.35pt}{1pt}{1pt}
\multicolumn{10}{l}{\textbf{Held-out Graph Node Classification}} \\
Computers
& $87.42 \pm 2.24$ & $79.81 \pm 0.54$ & $83.23 \pm 0.03$
& $75.04 \pm 0.19$ & $86.38 \pm 0.36$ & $\underline{87.66 \pm 0.33}$ & $78.57 \pm 0.16$ & $83.91 \pm 1.01$
& $\mathbf{88.85 \pm 0.46}$ \\
Physics
& $85.97 \pm 0.73$ & $92.36 \pm 0.08$ & $90.35 \pm 0.28$
& $92.34 \pm 0.11$ & $85.55 \pm 1.13$ & $\underline{95.36 \pm 0.13}$ & $90.95 \pm 0.56$ & $93.46 \pm 0.34$
& $\mathbf{95.83 \pm 0.06}$ \\
Citeseer
& $69.17 \pm 1.41$ & $71.39 \pm 0.76$ & $55.09 \pm 1.71$
& $\underline{71.56 \pm 0.89}$ & $68.94 \pm 1.51$ & $70.18 \pm 0.99$ & $69.16 \pm 0.37$ & $62.80 \pm 1.71$
& $\mathbf{72.81 \pm 0.73}$ \\
Pubmed
& $77.85 \pm 1.83$ & $83.11 \pm 0.59$ & $74.25 \pm 0.21$
& $81.73 \pm 0.22$ & $84.56 \pm 0.22$ & $\underline{85.34 \pm 0.28}$ & $79.07 \pm 1.20$ & $81.48 \pm 0.31$
& $\mathbf{87.41 \pm 0.18}$ \\
Cornell
& $47.03 \pm 3.03$ & $48.49 \pm 2.05$ & $44.60 \pm 1.63$
& $45.89 \pm 14.54$ & $\underline{54.20 \pm 3.28}$ & $42.65 \pm 7.85$ & $49.54 \pm 4.89$ & $44.22 \pm 1.67$
& $\mathbf{61.24 \pm 2.81}$ \\
Wisconsin
& $46.87 \pm 0.20$ & $56.12 \pm 1.59$ & $43.28 \pm 2.84$
& $52.04 \pm 11.10$ & $\underline{61.99 \pm 2.49}$ & $47.51 \pm 6.69$ & $49.47 \pm 2.58$ & $46.76 \pm 2.26$
& $\mathbf{68.75 \pm 2.50}$ \\
Arxiv
& $59.20 \pm 0.45$ & $57.19 \pm 0.53$ & $\underline{66.80 \pm 0.62}$
& $65.24 \pm 0.43$ & $59.64 \pm 0.52$ & $16.14 \pm 0.04$ & $61.32 \pm 0.77$ & $64.14 \pm 0.49$
& $\mathbf{70.21 \pm 0.71}$ \\
Products
& $77.16 \pm 0.87$ & $\underline{81.07 \pm 0.91}$ & $77.68 \pm 0.64$
& $58.58 \pm 0.37$ & $79.49 \pm 0.68$ & $27.31 \pm 0.02$ & $79.10 \pm 0.82$ & $80.88 \pm 0.96$
& $\mathbf{81.89 \pm 1.14}$ \\

\specialrule{0.35pt}{1pt}{1pt}
\multicolumn{10}{l}{\textbf{Graph-Level Transfer}} \\
COLLAB
& $\underline{73.78 \pm 1.06}$ & $71.48 \pm 1.13$ & $72.29 \pm 1.14$
& $71.41 \pm 1.10$ & -- & $72.34 \pm 1.19$ & $71.26 \pm 1.25$ & $70.74 \pm 1.89$
& $\mathbf{93.73 \pm 0.61}$ \\
DD
& $65.15 \pm 2.55$ & $66.38 \pm 2.74$ & $64.84 \pm 1.33$
& $\underline{74.00 \pm 2.12}$ & -- & $66.38 \pm 1.71$ & $65.32 \pm 1.89$ & $67.36 \pm 1.35$
& $\mathbf{91.75 \pm 1.48}$ \\
ENZYMES
& $24.17 \pm 2.25$ & $23.83 \pm 1.43$ & $\underline{25.63 \pm 0.86}$
& $20.88 \pm 2.35$ & -- & $22.13 \pm 2.82$ & $24.08 \pm 1.63$ & $23.25 \pm 3.79$
& $\mathbf{49.29 \pm 4.62}$ \\
IMDB-B
& $67.67 \pm 3.98$ & $67.10 \pm 3.25$ & $67.22 \pm 2.80$
& $66.02 \pm 3.24$ & -- & $69.27 \pm 1.83$ & $\underline{69.60 \pm 1.37}$ & $65.55 \pm 2.13$
& $\mathbf{90.37 \pm 1.28}$ \\
PROTEINS
& $63.95 \pm 2.87$ & $69.11 \pm 0.74$ & $65.99 \pm 4.30$
& $\underline{72.53 \pm 0.61}$ & -- & $66.44 \pm 2.20$ & $68.53 \pm 1.13$ & $65.90 \pm 2.82$
& $\mathbf{89.16 \pm 1.76}$ \\

\bottomrule
\end{tabular}
}
\end{table*}

On the source graphs, SPR obtains the best results on Cora, Photo, and CS, although it does not outperform GraphAny and GCOPE on Chameleon. The more important observation is its behavior beyond the pre-training domains: SPR achieves the best performance on all eight unseen graphs, including graphs that differ substantially in scale, feature space, and topology from the source graphs. The gains on Cornell and Wisconsin are particularly pronounced, while consistent improvements are also observed on much larger graphs such as Arxiv and Products. The fact that SPR does not uniformly dominate every source graph, yet performs consistently strongly on unseen graphs, suggests that its advantage is not simply stronger fitting to the pre-training domains. Instead, the learned representation remains effective as graph characteristics change, supporting flexible and transferable graph cognition under substantial domain variation.

For graph classification, SPR again achieves the best performance on all five benchmarks, exceeding the strongest competing result by $16.63$--$23.66$ percentage points. This provides a complementary evaluation of representation reusability because the same pretrained model is evaluated under a different prediction granularity. Together with the unseen-node results, these findings show that the representations learned by SPR remain effective across different graph domains and downstream settings. They are also consistent with the proposed prototype geometry serving as a shared relational reference rather than being specialized to a particular source topology.

\Needspace{8\baselineskip}
\subsection{Ablation Study}

\begin{wraptable}{r}{0.50\textwidth}
\vspace{-6pt}
\centering
\caption{Ablation of the key components of SPR.}
\label{tab:ablation_selected}
\scriptsize
\setlength{\tabcolsep}{1.8pt}
\renewcommand{\arraystretch}{1.0}
\resizebox{0.50\textwidth}{!}{
\begin{tabular}{lcccc}
\toprule
Setting & Cora & Cornell & COLLAB & ENZYMES \\
\midrule
SPR
& $84.10 \pm 0.54$ & $61.24 \pm 2.81$
& $93.73 \pm 0.61$ & $49.29 \pm 4.62$ \\

w/o spectral
& $83.62 \pm 0.12$ & $49.79 \pm 2.90$
& $74.73 \pm 1.70$ & $40.21 \pm 3.21$ \\

adj\_only
& $80.31 \pm 1.68$ & $46.07 \pm 4.75$
& $83.18 \pm 0.78$ & $46.50 \pm 4.17$ \\
heat\_only
& $81.23 \pm 0.54$ & $43.04 \pm 3.67$
& $84.44 \pm 0.36$ & $49.08 \pm 4.55$ \\

w/o proto
& $80.64 \pm 1.45$ & $46.48 \pm 3.61$
& $92.08 \pm 0.27$ & $47.33 \pm 3.96$ \\

w/o adapt
& $81.18 \pm 1.52$ & $45.38 \pm 3.25$
& $92.77 \pm 0.78$ & $48.00 \pm 5.27$ \\
\bottomrule
\end{tabular}
}
\vspace{-4pt}
\end{wraptable}

Table~\ref{tab:ablation_selected} examines whether the two core components of SPR account for the transfer behavior observed above. Removing spectral pattern parsing consistently degrades performance, with particularly large drops of $11.45$ points on Cornell, $19.00$ points on COLLAB, and $9.08$ points on ENZYMES, compared with only a modest decrease on the source graph Cora. This contrast indicates that the spectral component is especially important when the model encounters graphs or prediction settings beyond the source domains, supporting its role in providing a flexible spectral representation that adapts to graph-specific spectral variation rather than merely improving source-graph fitting.

The relational component exhibits a complementary effect. Removing the prototype geometry causes a $14.76$-point decrease on Cornell, while removing graph-specific adaptation to the prototype space leads to a $15.86$-point drop. Both degradations are substantially larger than that observed on Cora. These results support two aspects of the proposed relational mechanism: the shared prototype geometry provides relational information learned across source graphs, and this geometry needs to be instantiated according to each graph rather than applied uniformly. Finally, both \texttt{adj\_only} and \texttt{heat\_only} underperform the full model, showing that local adjacency or graph-specific diffusion alone cannot recover the information introduced by spectral pattern parsing and the shared relational geometry. Overall, the ablations separately support the necessity of the flexible spectral mechanism and the cross-graph relational reference used in SPR.

\subsection{Model Analysis}

We further examine whether the learned spectral responses and prototype geometry behave as intended beyond their contribution to downstream accuracy.

\begin{wrapfigure}{r}{0.50\textwidth}
  \vspace{-0.6em}
  \centering
  \begin{minipage}[t]{0.48\linewidth}
    \centering
    \includegraphics[width=\linewidth]{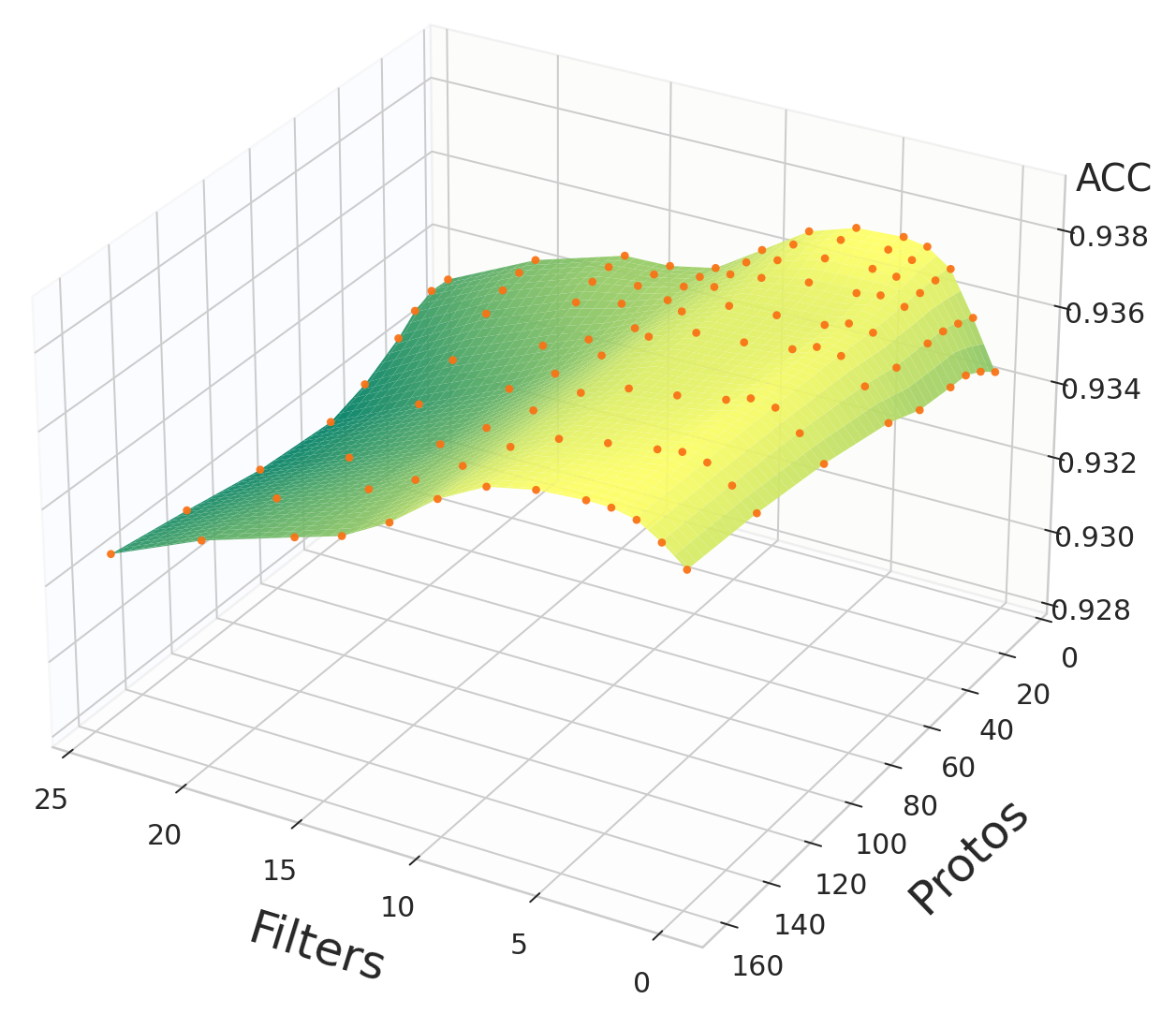}\\
    {\scriptsize (a) CS}
  \end{minipage}
  \hfill
  \begin{minipage}[t]{0.48\linewidth}
    \centering
    \includegraphics[width=\linewidth]{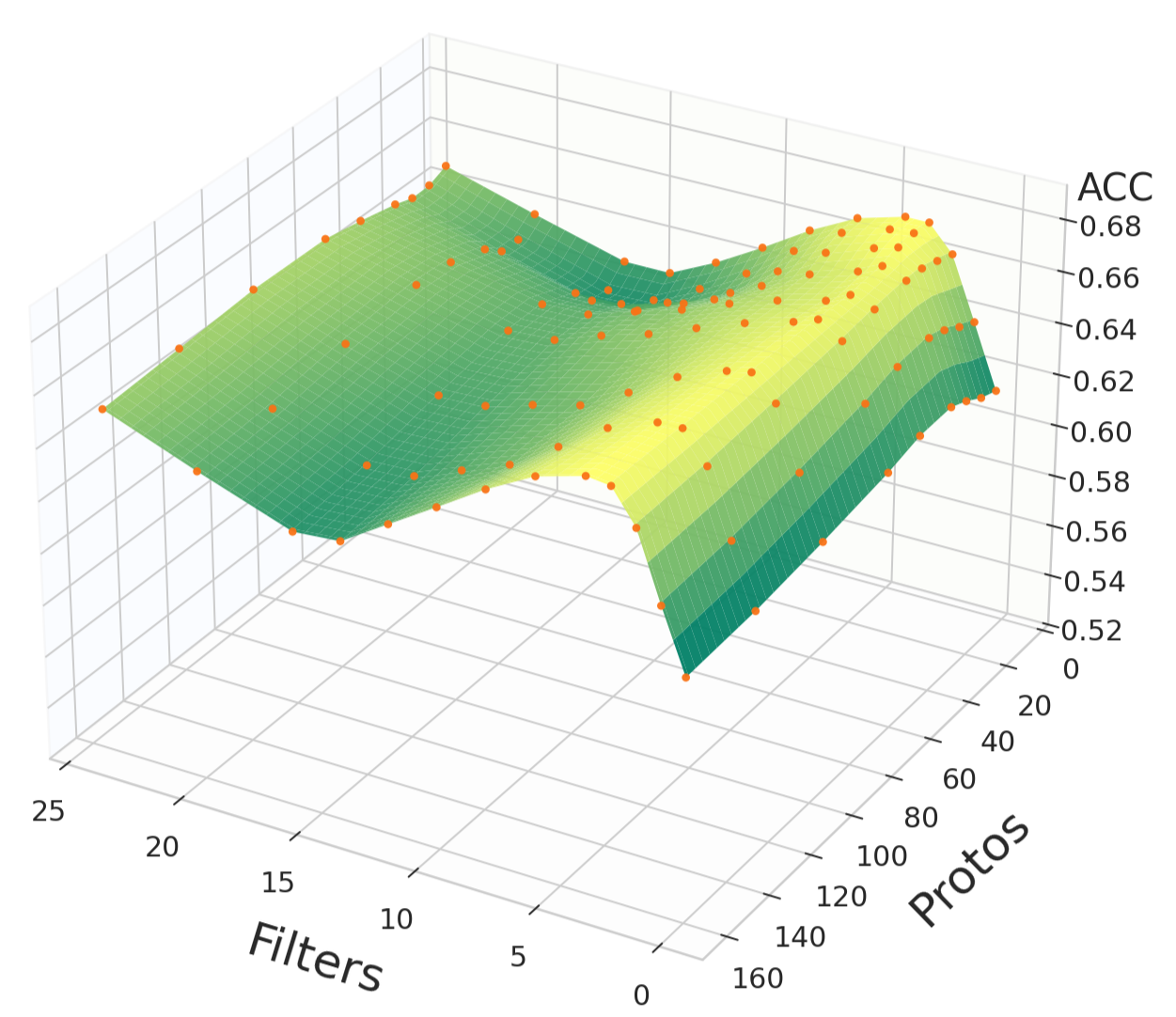}\\
    {\scriptsize (b) Wisconsin}
  \end{minipage}
  \caption{Effect of filters and prototypes on ACC.}
  \label{fig:surfaces}
  \vspace{-0.7em}
\end{wrapfigure}

As shown in Figure~\ref{fig:surfaces}, performance depends on both the number of spectral filters and the size of the prototype space. CS remains stable across most settings, whereas Wisconsin is more sensitive to the number of filters. Using too few filters degrades performance on both datasets, indicating that multiple spectral responses are needed to accommodate variation across spectral regions. In contrast, increasing the number of prototypes beyond a moderate size brings little additional benefit and can slightly reduce performance, suggesting that the recurring relational information can be summarized by a relatively compact prototype space.

\begin{wrapfigure}{r}{0.50\textwidth}
  \vspace{-0.5em}
  \centering
  \captionsetup[subfigure]{font=footnotesize,skip=1pt}
  \begin{subfigure}[t]{0.48\linewidth}
    \centering
    \includegraphics[width=\linewidth]{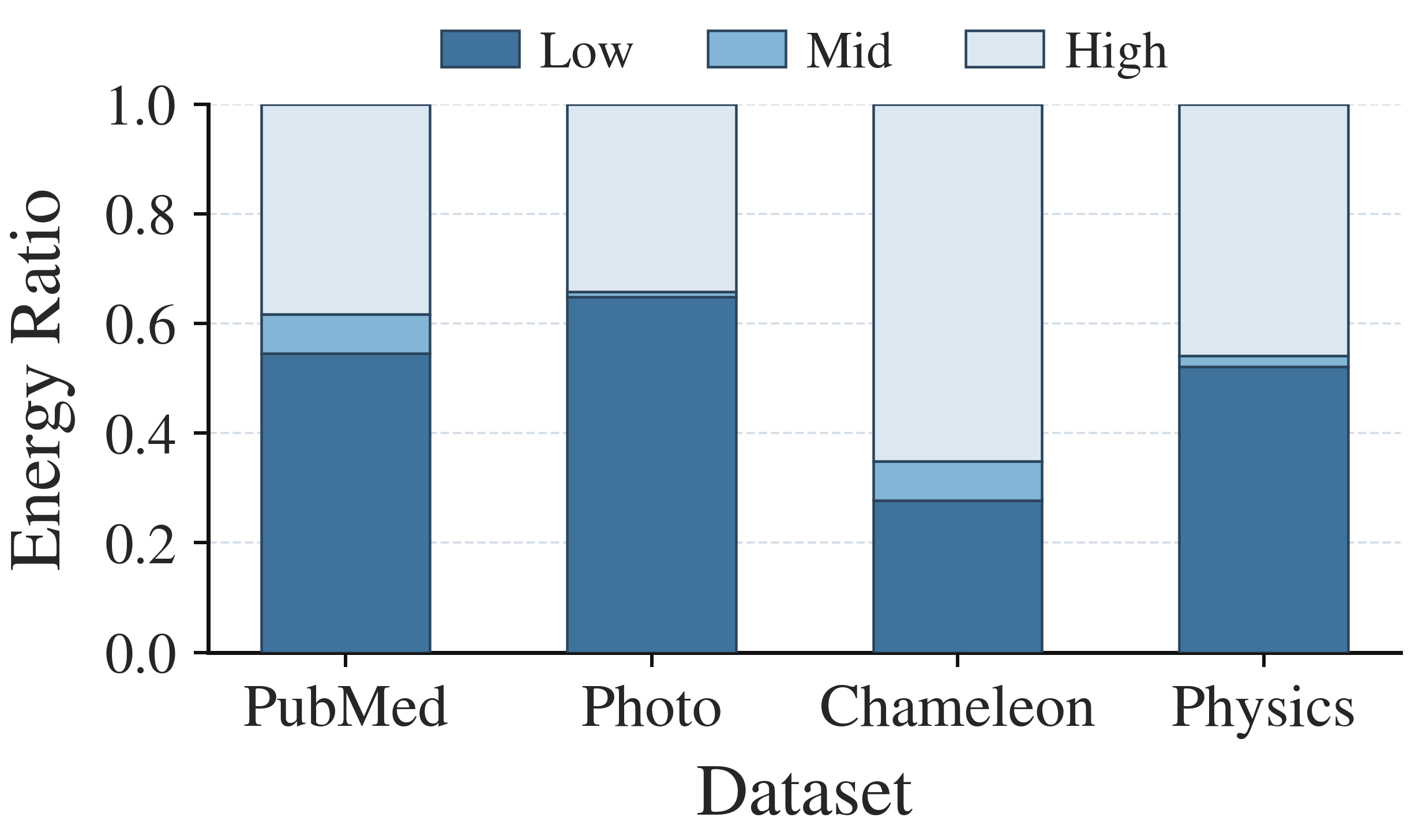}
    \caption{Spectral distribution.}
    \label{fig:xspec}
  \end{subfigure}
  \hfill
  \begin{subfigure}[t]{0.48\linewidth}
    \centering
    \includegraphics[width=\linewidth]{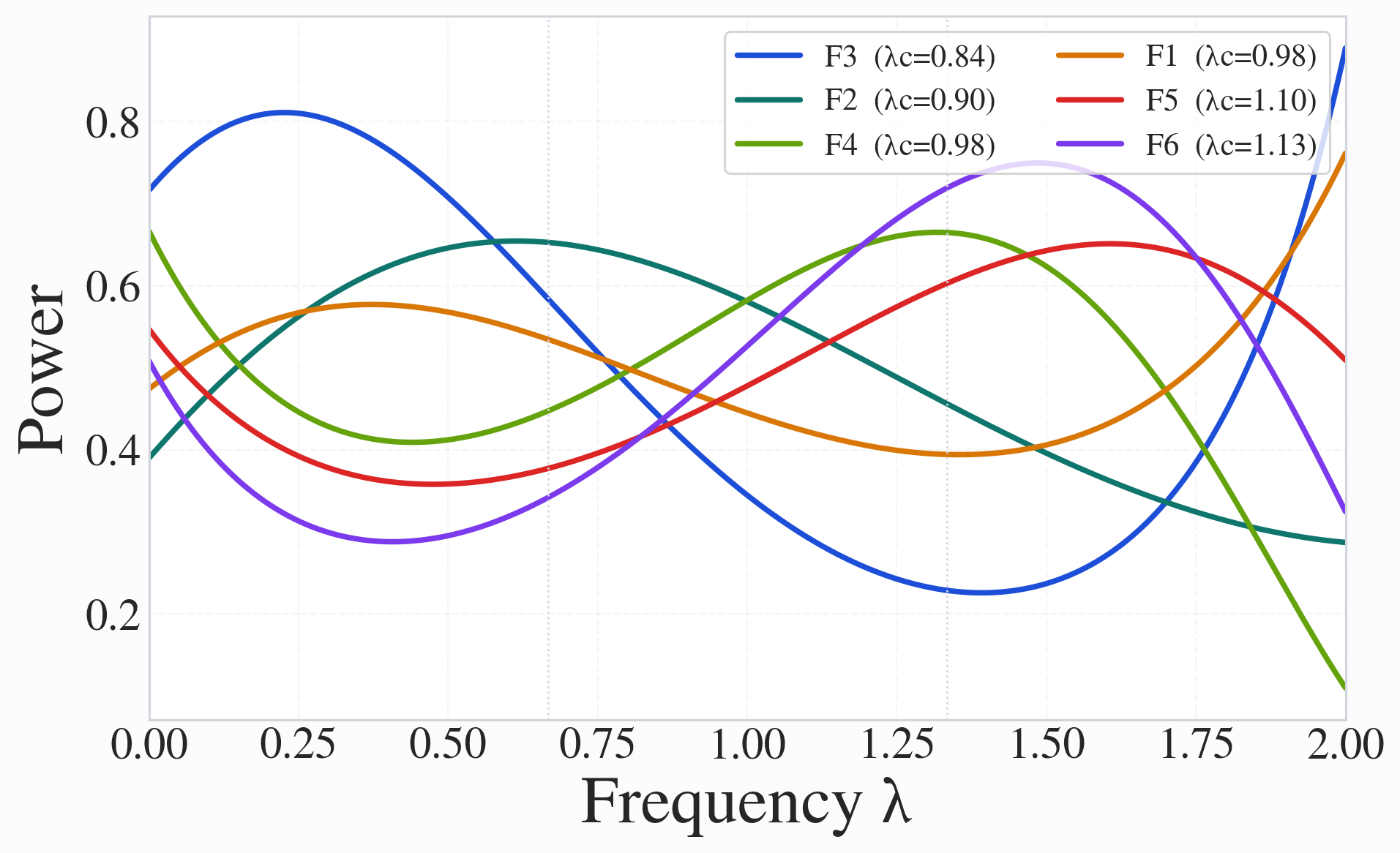}
    \caption{Spectral filter responses.}
    \label{fig:filter}
  \end{subfigure}
  \caption{Analysis of the learned spectral responses in SPR.}
  \label{fig:spectral_combined}
  \vspace{-0.7em}
\end{wrapfigure}

Figure~\ref{fig:spectral_combined} further examines what is learned by spectral pattern parsing. The response functions exhibit distinct peaks and power-weighted centers, showing that different filters specialize in complementary spectral regions rather than collapsing to similar responses. More importantly, the resulting spectral representations exhibit different energy distributions across graphs even though the response functions are shared. This behavior directly reflects the design of SPR: the spectral mechanism is shared across domains, while its realization remains graph-dependent through the graph-specific operator and input features. Compared with the raw spectral distributions in Figure~\ref{fig:raw_proxy_bands}, these results show that SPR adaptively reorganizes spectral information rather than imposing a fixed smoothing pattern.

\section{Conclusion}

This paper investigates transferable graph cognition from a spectral perspective. SPR learns shared Chebyshev spectral responses to flexibly characterize continuously varying graph patterns, while cross-graph relational geometry distills common pairwise organization from multiple source graphs into a shared prototype geometry. The shared spectral mechanism is instantiated on graph-specific operators, and the learned relational reference can be aligned to unseen graphs, together supporting graph cognition that is adaptive and transferable. Experiments across source and unseen graphs, node- and graph-level tasks, and few-shot settings support the effectiveness of the proposed design.

\subsection*{AI use statement}
Generative AI tools were used in a limited capacity for language polishing, minor rephrasing, improving manuscript organization and readability. Generative AI was not used to generate data or experimental results. All AI-assisted content was reviewed and verified by the authors, and all equations, citations, numerical results, technical descriptions, and scientific claims were independently checked by the authors. We take full responsibility for the final content of this work, including any text or artifacts produced with the aid of generative AI.

\bibliographystyle{plainnat}
\bibliography{1}

@inproceedings{DBLP:conf/icml/WangZMCZ025,
  author       = {Zehong Wang and
                  Zheyuan Zhang and
                  Tianyi Ma and
                  Nitesh V. Chawla and
                  Chuxu Zhang and
                  Yanfang Ye},
  editor       = {Aarti Singh and
                  Maryam Fazel and
                  Daniel Hsu and
                  Simon Lacoste{-}Julien and
                  Felix Berkenkamp and
                  Tegan Maharaj and
                  Kiri Wagstaff and
                  Jerry Zhu},
  title        = {Towards Graph Foundation Models: Learning Generalities Across Graphs
                  via Task-Trees},
  booktitle    = {Forty-second International Conference on Machine Learning, {ICML}
                  2025, Vancouver, BC, Canada, July 13-19, 2025},
  series       = {Proceedings of Machine Learning Research},
  volume       = {267},
  publisher    = {{PMLR} / OpenReview.net},
  year         = {2025},
  url          = {https://proceedings.mlr.press/v267/wang25eq.html},
  bibsource    = {dblp computer science bibliography, https://dblp.org}
}

@article{wang2025graph,
  title={Graph Foundation Models: A Comprehensive Survey},
  author={Wang, Zehong and Liu, Zheyuan and Ma, Tianyi and Li, Jiazheng and Zhang, Zheyuan and Fu, Xingbo and Li, Yiyang and Yuan, Zhengqing and Song, Wei and Ma, Yijun and others},
  journal={arXiv preprint arXiv:2505.15116},
  year={2025}
}

@inproceedings{he2025unigraph,
  title={Unigraph: Learning a unified cross-domain foundation model for text-attributed graphs},
  author={He, Yufei and Sui, Yuan and He, Xiaoxin and Hooi, Bryan},
  booktitle={Proceedings of the 31st ACM SIGKDD Conference on Knowledge Discovery and Data Mining V. 1},
  pages={448--459},
  year={2025}
}

@inproceedings{DBLP:conf/icml/0001H0GBB25,
  author       = {Nian Liu and
                  Xiaoxin He and
                  Thomas Laurent and
                  Francesco Di Giovanni and
                  Michael M. Bronstein and
                  Xavier Bresson},
  editor       = {Aarti Singh and
                  Maryam Fazel and
                  Daniel Hsu and
                  Simon Lacoste{-}Julien and
                  Felix Berkenkamp and
                  Tegan Maharaj and
                  Kiri Wagstaff and
                  Jerry Zhu},
  title        = {A General Graph Spectral Wavelet Convolution via Chebyshev Order Decomposition},
  booktitle    = {Forty-second International Conference on Machine Learning, {ICML}
                  2025, Vancouver, BC, Canada, July 13-19, 2025},
  series       = {Proceedings of Machine Learning Research},
  volume       = {267},
  publisher    = {{PMLR} / OpenReview.net},
  year         = {2025},
  url          = {https://proceedings.mlr.press/v267/liu25y.html},
  bibsource    = {dblp computer science bibliography, https://dblp.org}
}

@article{DBLP:journals/corr/abs-2510-01510,
  author       = {Jinwoo Kim and
                  Xingyue Huang and
                  Krzysztof Olejniczak and
                  Kyungbin Min and
                  Michael M. Bronstein and
                  Seunghoon Hong and
                  {.I}smail {.I}lkan Ceylan},
  title        = {Flock: {A} Knowledge Graph Foundation Model via Learning on Random
                  Walks},
  journal      = {CoRR},
  volume       = {abs/2510.01510},
  year         = {2025},
  url          = {https://doi.org/10.48550/arXiv.2510.01510},
  doi          = {10.48550/ARXIV.2510.01510},
  eprinttype   = {arXiv},
  eprint       = {2510.01510},
  bibsource    = {dblp computer science bibliography, https://dblp.org}
}

@article{DBLP:journals/corr/abs-2607-28980,
  author       = {Yi Wang and
                  Jitao Zhao and
                  Di Jin and
                  Dongxiao He},
  title        = {Beyond Feature and Structure Alignment: Learning Transferable Propagation
                  Knowledge for Graph Foundation Models},
  journal      = {CoRR},
  volume       = {abs/2607.28980},
  year         = {2026},
  url          = {https://doi.org/10.48550/arXiv.2607.28980},
  doi          = {10.48550/ARXIV.2607.28980},
  eprinttype   = {arXiv},
  eprint       = {2607.28980},
  bibsource    = {dblp computer science bibliography, https://dblp.org}
}

@inproceedings{DBLP:conf/www/0008HZW0Y25,
  author       = {Li Sun and
                  Zhenhao Huang and
                  Suyang Zhou and
                  Qiqi Wan and
                  Hao Peng and
                  Philip S. Yu},
  editor       = {Guodong Long and
                  Michale Blumestein and
                  Yi Chang and
                  Liane Lewin{-}Eytan and
                  Zi Helen Huang and
                  Elad Yom{-}Tov},
  title        = {RiemannGFM: Learning a Graph Foundation Model from Riemannian Geometry},
  booktitle    = {Proceedings of the {ACM} on Web Conference 2025, {WWW} 2025, Sydney,
                  NSW, Australia, 28 April 2025- 2 May 2025},
  pages        = {1154--1165},
  publisher    = {{ACM}},
  year         = {2025},
  url          = {https://doi.org/10.1145/3696410.3714952},
  doi          = {10.1145/3696410.3714952},
  bibsource    = {dblp computer science bibliography, https://dblp.org}
}

@inproceedings{DBLP:conf/kdd/PerozziAS14,
  author       = {Bryan Perozzi and
                  Rami Al{-}Rfou and
                  Steven Skiena},
  editor       = {Sofus A. Macskassy and
                  Claudia Perlich and
                  Jure Leskovec and
                  Wei Wang and
                  Rayid Ghani},
  title        = {DeepWalk: online learning of social representations},
  booktitle    = {The 20th {ACM} {SIGKDD} International Conference on Knowledge Discovery
                  and Data Mining, {KDD} '14, New York, NY, {USA} - August 24 - 27,
                  2014},
  pages        = {701--710},
  publisher    = {{ACM}},
  year         = {2014},
  url          = {https://doi.org/10.1145/2623330.2623732},
  doi          = {10.1145/2623330.2623732},
  bibsource    = {dblp computer science bibliography, https://dblp.org}
}

@inproceedings{DBLP:conf/kdd/YingHCEHL18,
  author       = {Rex Ying and
                  Ruining He and
                  Kaifeng Chen and
                  Pong Eksombatchai and
                  William L. Hamilton and
                  Jure Leskovec},
  editor       = {Yike Guo and
                  Faisal Farooq},
  title        = {Graph Convolutional Neural Networks for Web-Scale Recommender Systems},
  booktitle    = {Proceedings of the 24th {ACM} {SIGKDD} International Conference on
                  Knowledge Discovery {\&} Data Mining, {KDD} 2018, London, UK,
                  August 19-23, 2018},
  pages        = {974--983},
  publisher    = {{ACM}},
  year         = {2018},
  url          = {https://doi.org/10.1145/3219819.3219890},
  doi          = {10.1145/3219819.3219890},
  bibsource    = {dblp computer science bibliography, https://dblp.org}
}

@inproceedings{DBLP:conf/nips/FuLHDZPC25,
  author       = {Lele Fu and
                  Tianchi Liao and
                  Sheng Huang and
                  Bowen Deng and
                  Chuanfu Zhang and
                  Shirui Pan and
                  Chuan Chen},
  editor       = {Danielle Belgrave and
                  Cheng Zhang and
                  Laura N. Montoya and
                  Hsuan{-}Tien Lin and
                  Razvan Pascanu and
                  Piotr Koniusz and
                  Marzyeh Ghassemi and
                  Nancy Chen and
                  Iv{\'{a}}n Vladimir Meza Ru{\'{\i}}z and
                  Arturo Loaiza{-}Bonilla},
  title        = {Unsupervised Federated Graph Learning},
  booktitle    = {Advances in Neural Information Processing Systems 38: Annual Conference
                  on Neural Information Processing Systems 2025, NeurIPS 2025, San Diego,
                  CA, USA, December 2-7, 2025 / Mexico City, Mexico, November 30 - December
                  5, 2025},
  year         = {2025},
  url          = {http://papers.nips.cc/paper\_files/paper/2025/hash/8c4c80d3eee8a56da1302bc6dcd2d953-Abstract-Conference.html},
  bibsource    = {dblp computer science bibliography, https://dblp.org}
}

@inproceedings{DBLP:conf/nips/WangCMC25,
  author       = {Yuyao Wang and
                  Yu{-}Hung Cheng and
                  Debarghya Mukherjee and
                  Huimin Cheng},
  editor       = {Danielle Belgrave and
                  Cheng Zhang and
                  Laura N. Montoya and
                  Hsuan{-}Tien Lin and
                  Razvan Pascanu and
                  Piotr Koniusz and
                  Marzyeh Ghassemi and
                  Nancy Chen and
                  Iv{\'{a}}n Vladimir Meza Ru{\'{\i}}z and
                  Arturo Loaiza{-}Bonilla},
  title        = {Transfer Learning on Edge Connecting Probability Estimation Under
                  Graphon Model},
  booktitle    = {Advances in Neural Information Processing Systems 38: Annual Conference
                  on Neural Information Processing Systems 2025, NeurIPS 2025, San Diego,
                  CA, USA, December 2-7, 2025 / Mexico City, Mexico, November 30 - December
                  5, 2025},
  year         = {2025},
  url          = {http://papers.nips.cc/paper\_files/paper/2025/hash/30dee0ddc6c0b4907a9d587d3ce87979-Abstract-Conference.html},
  bibsource    = {dblp computer science bibliography, https://dblp.org}
}

@article{zhao2024fully,
  title={Fully-inductive node classification on arbitrary graphs},
  author={Zhao, Jianan and Zhu, Zhaocheng and Galkin, Mikhail and Mostafa, Hesham and Bronstein, Michael and Tang, Jian},
  journal={arXiv preprint arXiv:2405.20445},
  year={2024}
}

@article{wang2024gft,
  title={Gft: Graph foundation model with transferable tree vocabulary},
  author={Wang, Zehong and Zhang, Zheyuan and Chawla, Nitesh V and Zhang, Chuxu and Ye, Yanfang},
  journal={Advances in neural information processing systems},
  volume={37},
  pages={107403--107443},
  year={2024}
}

@inproceedings{OFA,
  author       = {Hao Liu and
                  Jiarui Feng and
                  Lecheng Kong and
                  Ningyue Liang and
                  Dacheng Tao and
                  Yixin Chen and
                  Muhan Zhang},
  title        = {One For All: Towards Training One Graph Model For All Classification
                  Tasks},
  booktitle    = {The Twelfth International Conference on Learning Representations,
                  {ICLR} 2024, Vienna, Austria, May 7-11, 2024},
  publisher    = {OpenReview.net},
  year         = {2024},
  bibsource    = {dblp computer science bibliography, https://dblp.org}
}

@article{DBLP:journals/corr/abs-2007-08663,
  author       = {Christopher Morris and
                  Nils M. Kriege and
                  Franka Bause and
                  Kristian Kersting and
                  Petra Mutzel and
                  Marion Neumann},
  title        = {TUDataset: {A} collection of benchmark datasets for learning with
                  graphs},
  journal      = {CoRR},
  volume       = {abs/2007.08663},
  year         = {2020},
  url          = {https://arxiv.org/abs/2007.08663},
  eprinttype   = {arXiv},
  eprint       = {2007.08663},
  bibsource    = {dblp computer science bibliography, https://dblp.org}
}

@article{hu2020open,
  title={Open graph benchmark: Datasets for machine learning on graphs},
  author={Hu, Weihua and Fey, Matthias and Zitnik, Marinka and Dong, Yuxiao and Ren, Hongyu and Liu, Bowen and Catasta, Michele and Leskovec, Jure},
  journal={Advances in neural information processing systems},
  volume={33},
  pages={22118--22133},
  year={2020}
}

@article{zhu2020deep,
  title={Deep graph contrastive representation learning},
  author={Zhu, Yanqiao and Xu, Yichen and Yu, Feng and Liu, Qiang and Wu, Shu and Wang, Liang},
  journal={arXiv preprint arXiv:2006.04131},
  year={2020}
}

@inproceedings{DBLP:conf/iclr/PeiWCLY20,
  author       = {Hongbin Pei and
                  Bingzhe Wei and
                  Kevin Chen{-}Chuan Chang and
                  Yu Lei and
                  Bo Yang},
  title        = {Geom-GCN: Geometric Graph Convolutional Networks},
  booktitle    = {8th International Conference on Learning Representations, {ICLR} 2020,
                  Addis Ababa, Ethiopia, April 26-30, 2020},
  publisher    = {OpenReview.net},
  year         = {2020},
  url          = {https://openreview.net/forum?id=S1e2agrFvS},
  bibsource    = {dblp computer science bibliography, https://dblp.org}
}

@article{DBLP:journals/corr/abs-1811-05868,
  author       = {Oleksandr Shchur and
                  Maximilian Mumme and
                  Aleksandar Bojchevski and
                  Stephan G{\"{u}}nnemann},
  title        = {Pitfalls of Graph Neural Network Evaluation},
  journal      = {CoRR},
  volume       = {abs/1811.05868},
  year         = {2018},
  url          = {http://arxiv.org/abs/1811.05868},
  eprinttype   = {arXiv},
  eprint       = {1811.05868},
  bibsource    = {dblp computer science bibliography, https://dblp.org}
}

@article{sen2008collective,
  title={Collective classification in network data},
  author={Sen, Prithviraj and Namata, Galileo and Bilgic, Mustafa and Getoor, Lise and Galligher, Brian and Eliassi-Rad, Tina},
  journal={AI magazine},
  volume={29},
  number={3},
  pages={93--93},
  year={2008}
}

@inproceedings{hou2023graphmae2,
  title={Graphmae2: A decoding-enhanced masked self-supervised graph learner},
  author={Hou, Zhenyu and He, Yufei and Cen, Yukuo and Liu, Xiao and Dong, Yuxiao and Kharlamov, Evgeny and Tang, Jie},
  booktitle={Proceedings of the ACM web conference 2023},
  pages={737--746},
  year={2023}
}

@inproceedings{DBLP:conf/www/ZhaoWLHJFZ26,
  author       = {Jitao Zhao and
                  Yi Wang and
                  Yawen Li and
                  Dongxiao He and
                  Di Jin and
                  Zhiyong Feng and
                  Weixiong Zhang},
  editor       = {Hakim Hacid and
                  Yoelle Maarek and
                  Francesco Bonchi and
                  Ido Guy and
                  Emine Yilmaz},
  title        = {Towards Graph Foundation Model: Node Feature Transfer Invariant Modeling
                  on General Graphs},
  booktitle    = {Proceedings of the {ACM} Web Conference 2026, {WWW} 2026, Dubai, United
                  Arab Emirates, originally scheduled for April 13-17, 2026, rescheduled
                  for June 29 - July 3, 2026},
  pages        = {810--821},
  publisher    = {{ACM}},
  year         = {2026},
  url          = {https://doi.org/10.1145/3774904.3792236},
  doi          = {10.1145/3774904.3792236},
  bibsource    = {dblp computer science bibliography, https://dblp.org}
}

@inproceedings{zhao2024all,
  title={All in one and one for all: A simple yet effective method towards cross-domain graph pretraining},
  author={Zhao, Haihong and Chen, Aochuan and Sun, Xiangguo and Cheng, Hong and Li, Jia},
  booktitle={Proceedings of the 30th ACM SIGKDD Conference on Knowledge Discovery and Data Mining},
  pages={4443--4454},
  year={2024}
}

@article{DBLP:journals/corr/abs-2605-08689,
  author       = {Xiaodong He and
                  Haolan He and
                  Ruiyi Fang and
                  Ming Sun and
                  Zhao Kang},
  title        = {Structure-Centric Graph Foundation Model via Geometric Bases},
  journal      = {CoRR},
  volume       = {abs/2605.08689},
  year         = {2026},
  url          = {https://doi.org/10.48550/arXiv.2605.08689},
  doi          = {10.48550/ARXIV.2605.08689},
  eprinttype   = {arXiv},
  eprint       = {2605.08689},
  bibsource    = {dblp computer science bibliography, https://dblp.org}
}

@article{DBLP:journals/corr/abs-2603-00618,
  author       = {Li Sun and
                  Zhenhao Huang and
                  Silei Chen and
                  Lanxu Yang and
                  Junda Ye and
                  Sen Su and
                  Philip S. Yu},
  title        = {Multi-Domain Riemannian Graph Gluing for Building Graph Foundation
                  Models},
  journal      = {CoRR},
  volume       = {abs/2603.00618},
  year         = {2026},
  url          = {https://doi.org/10.48550/arXiv.2603.00618},
  doi          = {10.48550/ARXIV.2603.00618},
  eprinttype   = {arXiv},
  eprint       = {2603.00618},
  bibsource    = {dblp computer science bibliography, https://dblp.org}
}

@inproceedings{yu2025samgpt,
  title={Samgpt: Text-free graph foundation model for multi-domain pre-training and cross-domain adaptation},
  author={Yu, Xingtong and Gong, Zechuan and Zhou, Chang and Fang, Yuan and Zhang, Hui},
  booktitle={Proceedings of the ACM on Web Conference 2025},
  pages={1142--1153},
  year={2025}
}

@inproceedings{DBLP:journals/corr/BrunaZSL13,
  author       = {Joan Bruna and
                  Wojciech Zaremba and
                  Arthur Szlam and
                  Yann LeCun},
  editor       = {Yoshua Bengio and
                  Yann LeCun},
  title        = {Spectral Networks and Locally Connected Networks on Graphs},
  booktitle    = {2nd International Conference on Learning Representations, {ICLR} 2014,
                  Banff, AB, Canada, April 14-16, 2014, Conference Track Proceedings},
  year         = {2014},
  url          = {http://arxiv.org/abs/1312.6203},
  bibsource    = {dblp computer science bibliography, https://dblp.org}
}

@inproceedings{titouan2019optimal,
  title={Optimal transport for structured data with application on graphs},
  author={Titouan, Vayer and Courty, Nicolas and Tavenard, Romain and Flamary, R{\'e}mi},
  booktitle={International Conference on Machine Learning},
  pages={6275--6284},
  year={2019},
  organization={PMLR}
}

@inproceedings{peyre2016gromov,
  title={Gromov-wasserstein averaging of kernel and distance matrices},
  author={Peyr{\'e}, Gabriel and Cuturi, Marco and Solomon, Justin},
  booktitle={International conference on machine learning},
  pages={2664--2672},
  year={2016},
  organization={PMLR}
}

@article{memoli2011gromov,
  title={Gromov--Wasserstein distances and the metric approach to object matching},
  author={M{\'e}moli, Facundo},
  journal={Foundations of computational mathematics},
  volume={11},
  number={4},
  pages={417--487},
  year={2011},
  publisher={Springer}
}

@article{defferrard2016convolutional,
  title={Convolutional neural networks on graphs with fast localized spectral filtering},
  author={Defferrard, Micha{\"e}l and Bresson, Xavier and Vandergheynst, Pierre},
  journal={Advances in neural information processing systems},
  volume={29},
  year={2016}
}

@inproceedings{DBLP:conf/iclr/DongSHWL25,
  author       = {Yushun Dong and
                  Patrick Soga and
                  Yinhan He and
                  Song Wang and
                  Jundong Li},
  title        = {Graph Neural Networks Are More Than Filters: Revisiting and Benchmarking
                  from {A} Spectral Perspective},
  booktitle    = {The Thirteenth International Conference on Learning Representations,
                  {ICLR} 2025, Singapore, April 24-28, 2025},
  publisher    = {OpenReview.net},
  year         = {2025},
  url          = {https://openreview.net/forum?id=nWdQX5hOL9},
  bibsource    = {dblp computer science bibliography, https://dblp.org}
}

@article{DBLP:journals/pieee/OrtegaFKMV18,
  author       = {Antonio Ortega and
                  Pascal Frossard and
                  Jelena Kovacevic and
                  Jos{\'{e}} M. F. Moura and
                  Pierre Vandergheynst},
  title        = {Graph Signal Processing: Overview, Challenges, and Applications},
  journal      = {Proc. {IEEE}},
  volume       = {106},
  number       = {5},
  pages        = {808--828},
  year         = {2018},
  url          = {https://doi.org/10.1109/JPROC.2018.2820126},
  doi          = {10.1109/JPROC.2018.2820126},
  bibsource    = {dblp computer science bibliography, https://dblp.org}
}

@article{DBLP:journals/focm/Memoli11,
  author       = {Facundo M{\'{e}}moli},
  title        = {Gromov-Wasserstein Distances and the Metric Approach to Object Matching},
  journal      = {Found. Comput. Math.},
  volume       = {11},
  number       = {4},
  pages        = {417--487},
  year         = {2011},
  url          = {https://doi.org/10.1007/s10208-011-9093-5},
  doi          = {10.1007/S10208-011-9093-5},
  bibsource    = {dblp computer science bibliography, https://dblp.org}
}

@inproceedings{DBLP:conf/iclr/KipfW17,
  author       = {Thomas N. Kipf and
                  Max Welling},
  title        = {Semi-Supervised Classification with Graph Convolutional Networks},
  booktitle    = {5th International Conference on Learning Representations, {ICLR} 2017,
                  Toulon, France, April 24-26, 2017, Conference Track Proceedings},
  publisher    = {OpenReview.net},
  year         = {2017},
  url          = {https://openreview.net/forum?id=SJU4ayYgl},
  bibsource    = {dblp computer science bibliography, https://dblp.org}
}

@article{zhang2023spectral,
  title={Spectral invariant learning for dynamic graphs under distribution shifts},
  author={Zhang, Zeyang and Wang, Xin and Zhang, Ziwei and Qin, Zhou and Wen, Weigao and Xue, Hui and Li, Haoyang and Zhu, Wenwu},
  journal={Advances in Neural Information Processing Systems},
  volume={36},
  pages={6619--6633},
  year={2023}
}

@inproceedings{chen2025autogfm,
  title={Autogfm: Automated graph foundation model with adaptive architecture customization},
  author={Chen, Haibo and Wang, Xin and Zhang, Zeyang and Li, Haoyang and Feng, Ling and Zhu, Wenwu},
  booktitle={Forty-second International Conference on Machine Learning},
  year={2025}
}

@article{Survey-GNNS,
  author       = {Zonghan Wu and
                  Shirui Pan and
                  Fengwen Chen and
                  Guodong Long and
                  Chengqi Zhang and
                  Philip S. Yu},
  title        = {A Comprehensive Survey on Graph Neural Networks},
  journal      = {{IEEE} Trans. Neural Networks Learn. Syst.},
  volume       = {32},
  number       = {1},
  pages        = {4--24},
  year         = {2021},
  doi          = {10.1109/TNNLS.2020.2978386},
  bibsource    = {dblp computer science bibliography, https://dblp.org}
}

@article{GNNsurvey2,
  title={Graph neural networks: A review of methods and applications},
  author={Zhou, Jie and Cui, Ganqu and Hu, Shengding and Zhang, Zhengyan and Yang, Cheng and Liu, Zhiyuan and Wang, Lifeng and Li, Changcheng and Sun, Maosong},
  journal={AI open},
  volume={1},
  pages={57--81},
  year={2020},
  publisher={Elsevier}
}

@inproceedings{GCN,
  author       = {Thomas N. Kipf and
                  Max Welling},
  title        = {Semi-Supervised Classification with Graph Convolutional Networks},
  booktitle    = {5th International Conference on Learning Representations, {ICLR} 2017,
                  Toulon, France, April 24-26, 2017, Conference Track Proceedings},
  publisher    = {OpenReview.net},
  year         = {2017},
  bibsource    = {dblp computer science bibliography, https://dblp.org}
}

@inproceedings{sun2025riemanngfm,
  title={Riemanngfm: Learning a graph foundation model from riemannian geometry},
  author={Sun, Li and Huang, Zhenhao and Zhou, Suyang and Wan, Qiqi and Peng, Hao and Yu, Philip},
  booktitle={Proceedings of the ACM on Web Conference 2025},
  pages={1154--1165},
  year={2025}
}

@inproceedings{DBLP:conf/www/LiuY0023,
  author       = {Zemin Liu and
                  Xingtong Yu and
                  Yuan Fang and
                  Xinming Zhang},
  editor       = {Ying Ding and
                  Jie Tang and
                  Juan F. Sequeda and
                  Lora Aroyo and
                  Carlos Castillo and
                  Geert{-}Jan Houben},
  title        = {GraphPrompt: Unifying Pre-Training and Downstream Tasks for Graph
                  Neural Networks},
  booktitle    = {Proceedings of the {ACM} Web Conference 2023, {WWW} 2023, Austin,
                  TX, USA, 30 April 2023 - 4 May 2023},
  pages        = {417--428},
  publisher    = {{ACM}},
  year         = {2023},
  url          = {https://doi.org/10.1145/3543507.3583386},
  doi          = {10.1145/3543507.3583386},
  bibsource    = {dblp computer science bibliography, https://dblp.org}
}

@inproceedings{DBLP:conf/www/ChenZWMLY26,
  author       = {Jialin Chen and
                  Haolan Zuo and
                  Haoyu Wang and
                  Siqi Miao and
                  Pan Li and
                  Rex Ying},
  editor       = {Hakim Hacid and
                  Yoelle Maarek and
                  Francesco Bonchi and
                  Ido Guy and
                  Emine Yilmaz},
  title        = {Towards {A} Universal Graph Structural Encoder},
  booktitle    = {Proceedings of the {ACM} Web Conference 2026, {WWW} 2026, Dubai, United
                  Arab Emirates, originally scheduled for April 13-17, 2026, rescheduled
                  for June 29 - July 3, 2026},
  pages        = {1457--1468},
  publisher    = {{ACM}},
  year         = {2026},
  url          = {https://doi.org/10.1145/3774904.3792656},
  doi          = {10.1145/3774904.3792656},
  bibsource    = {dblp computer science bibliography, https://dblp.org}
}

@article{liu2023one,
  title={One for all: Towards training one graph model for all classification tasks},
  author={Liu, Hao and Feng, Jiarui and Kong, Lecheng and Liang, Ningyue and Tao, Dacheng and Chen, Yixin and Zhang, Muhan},
  journal={arXiv preprint arXiv:2310.00149},
  year={2023}
}

@inproceedings{thakoor2021bootstrapped,
  title={Bootstrapped representation learning on graphs},
  author={Thakoor, Shantanu and Tallec, Corentin and Azar, Mohammad Gheshlaghi and Munos, R{\'e}mi and Veli{\v{c}}kovi{\'c}, Petar and Valko, Michal},
  booktitle={ICLR 2021 workshop on geometrical and topological representation learning},
  pages={1--14},
  year={2021},
  organization={OpenReview. net}
}

\appendix
\newpage
\section{Theory of Prototype Transfer}
\label{app:theory}

This section gives detailed proofs for two properties of prototype transfer. The first connects GW alignment to the fidelity of lifting the shared geometry back to a graph. The second studies the stability of the unnormalized prototype-induced operator $P_T=TK_\star T^\top$ with respect to transport estimation error. These results characterize complementary parts of the transfer mechanism rather than claiming an end-to-end guarantee for the normalized encoder operator.

For Proposition~\ref{prop:gw_lifting}, let \(T_m\in\Pi(w_m,b)\), so that
\[
T_m\mathbf{1}=w_m,
\qquad
T_m^\top\mathbf{1}=b,
\qquad
T_m(i,k)\geq 0.
\]
We assume \(w_m(i)>0\) on the node support considered below. Nodes with zero mass can be omitted because they contribute zero to the weighted error. We use
\[
\|B\|_{w_m\otimes w_m}^2
=
\sum_{i,i'}w_m(i)w_m(i')B(i,i')^2
\]
for the weighted squared Frobenius norm. For Proposition~\ref{prop:proto_propagation}, \(\|\cdot\|_2\) denotes the spectral norm and \(\|\cdot\|_F\) the Frobenius norm.

\subsection{Proof of Distortion-Controlled Geometry Transfer}
\label{app:proof_gw_lifting}

We prove that the same pairwise distortion minimized by the GW objective upper-bounds the error obtained after lifting the shared prototype geometry back to the graph.

\paragraph{Step 1: Normalize the transport into conditional prototype assignments.}
Define
\begin{equation}
R_m
=
\mathrm{diag}(w_m)^{-1}T_m .
\label{eq:app_row_normalized_transport}
\end{equation}
For any node \(i\) with \(w_m(i)>0\),
\begin{equation}
R_m(i,k)
=
\frac{T_m(i,k)}{w_m(i)}
\geq 0.
\label{eq:app_R_nonnegative}
\end{equation}
Moreover, because \(T_m\mathbf{1}=w_m\),
\begin{equation}
\begin{aligned}
\sum_k R_m(i,k)
&=
\frac{1}{w_m(i)}
\sum_k T_m(i,k) \\
&=
\frac{w_m(i)}{w_m(i)}
=
1.
\end{aligned}
\label{eq:app_R_rowsum}
\end{equation}
Therefore, each row \(R_m(i,\cdot)\) is a probability distribution over the \(p\) prototypes. It can be interpreted as the conditional prototype assignment of node \(i\).

\paragraph{Step 2: Express the lifted geometry as an expectation over prototype pairs.}
The lifted graph-space geometry is
\begin{equation}
\overline C_m
=
R_m C_\star R_m^\top .
\label{eq:app_lifted_geometry}
\end{equation}
Its \((i,i')\)-th entry is
\begin{equation}
\begin{aligned}
\overline C_m(i,i')
&=
e_i^\top R_m C_\star R_m^\top e_{i'} \\
&=
\sum_{k,l}
R_m(i,k)\,
C_\star(k,l)\,
R_m(i',l).
\end{aligned}
\label{eq:app_lifted_geometry_entry}
\end{equation}
For this fixed node pair, define the product distribution
\begin{equation}
\pi_{ii'}(k,l)
=
R_m(i,k)R_m(i',l).
\label{eq:app_product_assignment}
\end{equation}
Because both rows of \(R_m\) sum to one,
\begin{equation}
\sum_{k,l}\pi_{ii'}(k,l)
=
\left(\sum_kR_m(i,k)\right)
\left(\sum_lR_m(i',l)\right)
=
1.
\label{eq:app_product_assignment_sum}
\end{equation}
Hence, if \(K\sim R_m(i,\cdot)\) and \(L\sim R_m(i',\cdot)\) are sampled independently, then
\begin{equation}
\overline C_m(i,i')
=
\mathbb{E}\!\left[C_\star(K,L)\mid i,i'\right].
\label{eq:app_lifted_expectation}
\end{equation}

\paragraph{Step 3: Bound the reconstruction error for one node pair.}
Since \(C_m(i,i')\) is fixed once \((i,i')\) is fixed, Eq.~\eqref{eq:app_lifted_expectation} gives
\begin{equation}
\begin{aligned}
C_m(i,i')-\overline C_m(i,i')
&=
C_m(i,i')
-
\mathbb{E}\!\left[C_\star(K,L)\mid i,i'\right] \\
&=
\mathbb{E}\!\left[
C_m(i,i')-C_\star(K,L)
\mid i,i'
\right].
\end{aligned}
\label{eq:app_lifting_error_expectation}
\end{equation}
Applying Jensen's inequality to the convex function \(x\mapsto x^2\),
\begin{equation}
\begin{aligned}
\bigl(
C_m(i,i')-\overline C_m(i,i')
\bigr)^2
&=
\left(
\mathbb{E}\!\left[
C_m(i,i')-C_\star(K,L)
\mid i,i'
\right]
\right)^2 \\
&\leq
\mathbb{E}\!\left[
\bigl(
C_m(i,i')-C_\star(K,L)
\bigr)^2
\mid i,i'
\right].
\end{aligned}
\label{eq:app_jensen_step}
\end{equation}
Expanding this expectation under the product assignment in Eq.~\eqref{eq:app_product_assignment},
\begin{equation}
\begin{aligned}
\bigl(
C_m(i,i')-\overline C_m(i,i')
\bigr)^2
\leq
\sum_{k,l}
\bigl(
C_m(i,i')-C_\star(k,l)
\bigr)^2
R_m(i,k)R_m(i',l).
\end{aligned}
\label{eq:app_pairwise_lifting_bound}
\end{equation}
Thus, for each node pair, the squared error of the averaged lifted geometry is no larger than the expected squared relational mismatch before averaging.

\paragraph{Step 4: Aggregate over node pairs and recover the GW distortion.}
Multiply Eq.~\eqref{eq:app_pairwise_lifting_bound} by
\(w_m(i)w_m(i')\) and sum over all node pairs:
\begin{equation}
\begin{aligned}
\|C_m-\overline C_m\|_{w_m\otimes w_m}^2
&=
\sum_{i,i'}
w_m(i)w_m(i')
\bigl(
C_m(i,i')-\overline C_m(i,i')
\bigr)^2 \\
&\leq
\sum_{i,i',k,l}
w_m(i)w_m(i')
\bigl(
C_m(i,i')-C_\star(k,l)
\bigr)^2
R_m(i,k)R_m(i',l).
\end{aligned}
\label{eq:app_weighted_lifting_1}
\end{equation}
From the definition of \(R_m\),
\begin{equation}
w_m(i)R_m(i,k)
=
T_m(i,k),
\qquad
w_m(i')R_m(i',l)
=
T_m(i',l).
\label{eq:app_mass_recovery}
\end{equation}
Substituting Eq.~\eqref{eq:app_mass_recovery} into Eq.~\eqref{eq:app_weighted_lifting_1},
\begin{equation}
\begin{aligned}
\|C_m-\overline C_m\|_{w_m\otimes w_m}^2
&\leq
\sum_{i,i',k,l}
\bigl(
C_m(i,i')-C_\star(k,l)
\bigr)^2
T_m(i,k)T_m(i',l) \\
&=
\mathcal{D}_{\rm GW}(C_m,C_\star,T_m).
\end{aligned}
\label{eq:app_gw_lifting_final}
\end{equation}
The last line is the GW relational distortion at \(T_m\), which proves Proposition~\ref{prop:gw_lifting}.

\begin{proposition}[Transport-stable prototype propagation]
\label{prop:proto_propagation}
Let $P_T=TK_\star T^\top$ and $P_{\widetilde T}=\widetilde T K_\star\widetilde T^\top$. Then
\begin{equation}
\|P_T-P_{\widetilde T}\|_F
\leq \|K_\star\|_2(\|T\|_2+\|\widetilde T\|_2)\|T-\widetilde T\|_F.
\end{equation}
If $T$ and $\widetilde T$ are probability couplings, the bound reduces to $2\|K_\star\|_2\|T-\widetilde T\|_F$.
\end{proposition}
The result applies to the unnormalized operator before the symmetric normalization used by the encoder.

\subsection{Proof of Transport-Stable Prototype Propagation}
\label{app:proof_proto_propagation}

We next bound the change in the induced node-space operator under perturbations of the graph-to-prototype transport.

Let
\begin{equation}
P_T
=
T K_\star T^\top,
\qquad
P_{\widetilde T}
=
\widetilde T K_\star \widetilde T^\top,
\label{eq:app_two_operators}
\end{equation}
and define the transport perturbation
\begin{equation}
\Delta_T
=
T-\widetilde T.
\label{eq:app_transport_delta}
\end{equation}

\begin{equation}
\begin{aligned}
P_T-P_{\widetilde T}
&=
T K_\star T^\top
-
\widetilde T K_\star \widetilde T^\top .
\end{aligned}
\label{eq:app_operator_difference_start}
\end{equation}
Add and subtract \(\widetilde T K_\star T^\top\):
\begin{equation}
\begin{aligned}
P_T-P_{\widetilde T}
&=
T K_\star T^\top
-
\widetilde T K_\star T^\top
+
\widetilde T K_\star T^\top
-
\widetilde T K_\star \widetilde T^\top \\
&=
(T-\widetilde T)K_\star T^\top
+
\widetilde T K_\star
(T-\widetilde T)^\top \\
&=
\Delta_T K_\star T^\top
+
\widetilde T K_\star\Delta_T^\top .
\end{aligned}
\label{eq:app_operator_difference_decomposition}
\end{equation}

We use the standard inequalities
\[
\|AB\|_F\leq\|A\|_F\|B\|_2,
\qquad
\|AB\|_F\leq\|A\|_2\|B\|_F.
\]
For the first term in Eq.~\eqref{eq:app_operator_difference_decomposition},
\begin{equation}
\begin{aligned}
\|\Delta_T K_\star T^\top\|_F
&\leq
\|\Delta_T\|_F
\|K_\star T^\top\|_2 \\
&\leq
\|\Delta_T\|_F
\|K_\star\|_2
\|T^\top\|_2 \\
&=
\|K_\star\|_2
\|T\|_2
\|\Delta_T\|_F .
\end{aligned}
\label{eq:app_first_transport_term}
\end{equation}

Similarly,
\begin{equation}
\begin{aligned}
\|\widetilde T K_\star\Delta_T^\top\|_F
&\leq
\|\widetilde T K_\star\|_2
\|\Delta_T^\top\|_F \\
&\leq
\|\widetilde T\|_2
\|K_\star\|_2
\|\Delta_T\|_F,
\end{aligned}
\label{eq:app_second_transport_term}
\end{equation}
where we used
\(\|\Delta_T^\top\|_F=\|\Delta_T\|_F\).

Using the triangle inequality and Eqs.~\eqref{eq:app_first_transport_term}--\eqref{eq:app_second_transport_term},
\begin{equation}
\begin{aligned}
\|P_T-P_{\widetilde T}\|_F
&\leq
\|\Delta_T K_\star T^\top\|_F
+
\|\widetilde T K_\star\Delta_T^\top\|_F \\
&\leq
\|K_\star\|_2
\|T\|_2
\|\Delta_T\|_F
+
\|\widetilde T\|_2
\|K_\star\|_2
\|\Delta_T\|_F \\
&=
\|K_\star\|_2
\bigl(
\|T\|_2+\|\widetilde T\|_2
\bigr)
\|T-\widetilde T\|_F .
\end{aligned}
\label{eq:app_transport_stability_general}
\end{equation}

Suppose \(T\) is a probability coupling, i.e., \(T(i,k)\geq0\) and
\begin{equation}
\sum_{i,k}T(i,k)=1.
\label{eq:app_transport_total_mass}
\end{equation}
Then
\begin{equation}
\begin{aligned}
\|T\|_2
&\leq
\|T\|_F \\
&=
\left(\sum_{i,k}T(i,k)^2\right)^{1/2} \\
&\leq
\sum_{i,k}T(i,k) \\
&=
1.
\end{aligned}
\label{eq:app_probability_transport_norm}
\end{equation}
The same argument gives
\(\|\widetilde T\|_2\leq1\).
Substituting these two inequalities into
Eq.~\eqref{eq:app_transport_stability_general},
\begin{equation}
\boxed{
\|P_T-P_{\widetilde T}\|_F
\leq
2\|K_\star\|_2
\|T-\widetilde T\|_F
}.
\label{eq:app_transport_stability_probability}
\end{equation}
This proves the second claim.

\paragraph{Consequence for the propagated representation.}
For a fixed node representation \(H\), the operator perturbation produces
\begin{equation}
\begin{aligned}
\|(P_T-P_{\widetilde T})H\|_F
&\leq
\|P_T-P_{\widetilde T}\|_F
\|H\|_2.
\end{aligned}
\label{eq:app_representation_consequence_1}
\end{equation}
Combining this with Eq.~\eqref{eq:app_transport_stability_general} gives
\begin{equation}
\begin{aligned}
\|(P_T-P_{\widetilde T})H\|_F
&\leq
\|K_\star\|_2
\bigl(
\|T\|_2+\|\widetilde T\|_2
\bigr)
\|T-\widetilde T\|_F
\|H\|_2.
\end{aligned}
\label{eq:app_representation_consequence_2}
\end{equation}
Thus, transport perturbations also bound their effect on a fixed propagated representation.

\section{Notations Used in Our Method}

Table~\ref{tab:notation} summarizes the main notation used by SPR. Unless otherwise specified, all graphs are treated as undirected during propagation, and sparse matrices are used for adjacency-based operations.

\begin{table*}[htbp]
\centering
\caption{Notation used throughout the method and appendix.}
\label{tab:notation}
\small
\setlength{\tabcolsep}{5pt}
\renewcommand{\arraystretch}{1.12}
\begin{tabular}{lll}
\toprule
Symbol & Shape / Type & Description \\
\midrule
$G_m$ & graph & The $m$-th graph domain. \\
$V_m, E_m$ & node / edge set & Node set and edge set of $G_m$. \\
$n_m, e_m$ & scalar & Number of nodes and edges in $G_m$. \\
$A_m$ & $n_m\times n_m$ & Adjacency matrix of $G_m$. \\
$D_m$ & $n_m\times n_m$ & Degree matrix of $A_m$. \\
$L_m$ & $n_m\times n_m$ & Normalized graph Laplacian. \\
$S_m$ & $n_m\times n_m$ & Normalized propagation operator used by the spectral frontend. \\
$\rho_m$ & $n_m$ & Propagation-domain eigenvalues of $S_m$; $\rho=1-\lambda$ for Laplacian frequency $\lambda$. \\
$X_m$ & $n_m\times d_m$ & Raw node feature matrix, where $d_m$ may vary across domains. \\
$M_f$ & scalar & Number of learnable Chebyshev spectral filters. \\
$Z_\alpha$ & $n_m\times d_m$ & Filtered response produced by the $\alpha$-th Chebyshev spectral filter. \\
$K_c$ & scalar & Chebyshev polynomial order. \\
$r$ & scalar & Truncated SVD rank for each spectral filter response. \\
$d$ & scalar & Unified spectral feature dimension after parsing. \\
$h$ & scalar & Hidden dimension of the propagation backbone. \\
$d_z$ & scalar & Output embedding dimension. \\
$C_m$ & $n_m\times n_m$ & Heat-kernel-distance structural cost matrix for graph $G_m$. \\
$p$ & scalar & Number of structural prototypes. \\
$C_\star$ & $p\times p$ & Shared GW barycenter cost matrix in prototype space. \\
$A_\star$ & $p\times p$ & Prototype affinity matrix derived from $C_\star$. \\
$K_\star$ & $p\times p$ & Prototype heat kernel derived from $C_\star$. \\
$T_m$ & $n_m\times p$ & Node-to-prototype transport plan for graph $G_m$. \\
$w_m$ & $n_m$ & Node mass distribution of graph $G_m$. \\
$b$ & $p$ & Prototype mass distribution. \\
$\lambda_m$ & scalar & Weight of source graph $G_m$ in the GW barycenter. \\
$P_m$ & $n_m\times n_m$ & Prototype-induced propagation operator on graph $G_m$. \\
$K_{t_\ell}^{(m)}$ & $n_m\times n_m$ & Heat-diffusion operator at layer $\ell$. \\
$H^{(\ell)}$ & $n_m\times h$ & Node representation after layer $\ell$. \\
$H_A^{(\ell)}$ & $n_m\times h$ & Local adjacency-path representation. \\
$H_K^{(\ell)}$ & $n_m\times h$ & Heat-diffusion-path representation. \\
$H_P^{(\ell)}$ & $n_m\times h$ & Prototype-path representation. \\
$g_i^{(\ell)}$ & $\mathbb{R}^{3}$ & Node-wise gate over the three propagation paths. \\
$\mathcal{M}$ & set of nodes & Masked node set for spectral-feature reconstruction. \\
\bottomrule
\end{tabular}
\end{table*}

\section{Algorithm}
\begin{algorithm}[H]
  \caption{Spectral Prototype-Guided Pre-Training}
  \label{alg:spectral_proto_pretrain}
  \begin{center}
    \begin{minipage}{0.95\columnwidth}
      \begin{algorithmic}[1]
      \normalsize

      \STATE \textbf{Input:} Source graphs $\{G_m=(V_m,E_m,X_m)\}_{m=1}^{M}$, domain weights $\{\lambda_m\}_{m=1}^{M}$;
      \STATE \hspace{1em} prototype number $p$, filter number $M_f$, Chebyshev order $K_c$, SVD rank $r$;
      \STATE \hspace{1em} layers $L$, heat times $\{t_\ell\}_{\ell=1}^{L}$.
      \STATE \textbf{Output:} Pretrained encoder $f_\theta$, shared prototype kernel $K_\star$.

      \STATE \textbf{(1) Shared prototype geometry}
      \FOR{each source graph $G_m$}
        \STATE $L_m \leftarrow I-D_m^{-\frac{1}{2}}A_mD_m^{-\frac{1}{2}}$
        \STATE $K_{t_{\mathrm{gw}}}^{(m)} \leftarrow \exp(-t_{\mathrm{gw}}L_m)$
        \FOR{$\ell=1$ to $L$}
          \STATE $K_{t_\ell}^{(m)} \leftarrow \exp(-t_\ell L_m)$
        \ENDFOR
        \STATE $C_m \leftarrow \mathrm{HeatKernelCost}(K_{t_{\mathrm{gw}}}^{(m)})$
      \ENDFOR
      \STATE $C_\star,\{T_m\}_{m=1}^{M} \leftarrow \mathrm{GWBarycenter}(\{C_m\}_{m=1}^{M},\{\lambda_m\}_{m=1}^{M},p)$
      \STATE $A_\star \leftarrow \mathrm{Affinity}(C_\star)$
      \STATE $K_\star \leftarrow \exp(-L(A_\star))$

      \WHILE{not converged}
        \FOR{each source graph $G_m$}

          \STATE \textbf{(2) Spectral frontend}
          \STATE $S_m \leftarrow D_m^{-\frac{1}{2}}A_mD_m^{-\frac{1}{2}}$
          \FOR{$\alpha=1$ to $M_f$}
            \STATE $Z_\alpha \leftarrow \sum_{k=0}^{K_c}\theta_{\alpha,k}T_k(S_m,X_m)$
            \STATE $\widetilde{Z}_\alpha \leftarrow \mathrm{SVD}^{\mathrm{node}}_r(Z_\alpha)$
          \ENDFOR
          \STATE $X_{\mathrm{spec},m} \leftarrow \mathrm{Proj}([\widetilde{Z}_1\Vert\cdots\Vert\widetilde{Z}_{M_f}])$

          \STATE \textbf{(3) Prototype-guided graph propagation}
          \STATE $P_m \leftarrow \mathrm{SymNorm}(T_mK_\star T_m^\top)$
          \STATE sample masked node set $\mathcal{M}$ and set $\widehat X_{\mathrm{spec},m}\leftarrow\mathrm{Mask}(X_{\mathrm{spec},m},\mathcal{M})$
          \STATE $H_m^{(0)} \leftarrow \mathrm{InputProj}(\widehat X_{\mathrm{spec},m})$
          \STATE $H_{A,m}^{(0)},H_{K,m}^{(0)},H_{P,m}^{(0)} \leftarrow H_m^{(0)},H_m^{(0)},H_m^{(0)}$

          \FOR{$\ell=1$ to $L$}
            \STATE $H_{A,m}^{(\ell)} \leftarrow \mathrm{GCN}_{A}^{(\ell)}(A_m,H_{A,m}^{(\ell-1)})$
            \STATE $H_{K,m}^{(\ell)} \leftarrow \mathrm{GCN}_{K}^{(\ell)}(K_{t_\ell}^{(m)},H_{K,m}^{(\ell-1)})$
            \STATE $H_{P,m}^{(\ell)} \leftarrow \mathrm{GCN}_{P}^{(\ell)}(P_m,H_{P,m}^{(\ell-1)})$

            \STATE $[g_{A,m}^{(\ell)},g_{K,m}^{(\ell)},g_{P,m}^{(\ell)}]
            \leftarrow
            \mathrm{Softmax}(\mathrm{Gate}([H_{A,m}^{(\ell)}\Vert H_{K,m}^{(\ell)}\Vert H_{P,m}^{(\ell)}]))$

            \STATE $H_m^{(\ell)} \leftarrow
            g_{A,m}^{(\ell)}\odot H_{A,m}^{(\ell)}
            +g_{K,m}^{(\ell)}\odot H_{K,m}^{(\ell)}
            +g_{P,m}^{(\ell)}\odot H_{P,m}^{(\ell)}$

            \STATE $H_{A,m}^{(\ell)},H_{K,m}^{(\ell)},H_{P,m}^{(\ell)}
            \leftarrow H_m^{(\ell)},H_m^{(\ell)},H_m^{(\ell)}$
          \ENDFOR

          \STATE $Z_m \leftarrow H_m^{(L)}$

          \STATE \textbf{(4) Self-supervised losses}
          \STATE sample a second edge-perturbed view and obtain $Z'_m$ using the same spectral frontend and encoder
          \STATE $\mathcal{L}_{\mathrm{node}} \leftarrow \mathrm{InfoNCE}(Z_m,Z'_m)$
          \STATE $\mathcal{L}_{\mathrm{spec}} \leftarrow |\mathcal{M}|^{-1}\sum_{i\in\mathcal{M}}\|f_{\mathrm{rec}}(Z_{m,i})-X_{\mathrm{spec},m,i}\|_2^2$

          \STATE $\mathcal{L}_m \leftarrow
          1.5\mathcal{L}_{\mathrm{node}}
          +0.3\mathcal{L}_{\mathrm{spec}}$

          \STATE Update the trainable encoder and spectral parameters by minimizing $\mathcal{L}_m$.
        \ENDFOR
      \ENDWHILE

      \STATE \textbf{return} $f_\theta$ and $K_\star$
      \end{algorithmic}
    \end{minipage}
  \end{center}
\end{algorithm}
Algorithm~\ref{alg:spectral_proto_pretrain} summarizes SPR pre-training. It first learns a shared GW prototype geometry and converts it into \(K_\star\). 
Each source graph is then encoded by spectral parsing, followed by gated propagation over adjacency, heat diffusion, and the symmetrically normalized prototype-induced operator derived from \(T_mK_\star T_m^\top\). 
The encoder is trained with node-level contrastive learning and spectral feature reconstruction.

\section{Implementation Details}
\label{app:impl}
\begin{table}[H]
\centering
\caption{Main hyper-parameters.}
\label{tab:impl_hparams}
\small
\setlength{\tabcolsep}{5pt}
\renewcommand{\arraystretch}{1.12}
\resizebox{\textwidth}{!}{%
\begin{tabular}{lll}
\toprule
Category & Hyper-parameter & Value \\
\midrule
Pre-training domains & Source graphs & Cora, CS, Photo, Chameleon \\
Backbone & epoch & $25$ \\
Backbone & Input / hidden / output dimensions & $128/1024/768$ \\
Backbone & GCN layers & $3$ \\
Backbone & Dropout & $0.12$ \\
Backbone & Heat times & $[0.5,1.0,1.5]$ \\
Spectral frontend & Number of Chebyshev filters $M_f$ & $6$ \\
Spectral frontend & Chebyshev order $K_c$ & $3$ \\
Spectral frontend & SVD rank $r$ & $32$ \\
Spectral frontend & projected dimensions for input & $128$ \\
Prototype module & Number of prototypes $p$ & $96$ \\
Prototype module & GW diffusion time $t_{\mathrm{gw}}$ & $0.58$ \\
Prototype module & GW entropy regularization $\epsilon_{\mathrm{gw}}$ & $0.005$ \\
Optimization & Optimizer & Adam \\
Optimization & Learning rate & $1.2\times10^{-4}$ \\
Optimization & Weight decay & $10^{-5}$ \\
Loss & Loss weights & $1.5/0.3$ for node/feature\\
Downstream adaptation & K-means assignment temperature & $0.2$ \\
Downstream adaptation & sparse prototype assignment & top-$8$ \\
Downstream adaptation & Linear probing epochs & $10000$ \\
Downstream adaptation & Linear probing learning rate & $0.015$ \\
\bottomrule
\end{tabular}}
\end{table}

\paragraph{Baselines.}
The baselines cover self-supervised graph representation learning and graph foundation or transfer learning methods, including GraphMAE2, BGRL, GRACE, GCOPE, GraphAny, TIG, SAMGPT, and GraphGlue. We use released implementations and recommended settings for baseline-specific pre-training or adaptation, while evaluating all supported methods with the same downstream splits, shot construction, and accuracy metric described below. Datasets unsupported by a baseline are marked as missing.

\paragraph{Data splits and evaluation protocol.}
For node classification, we use stratified $1{:}1{:}8$ train/validation/test splits and report mean and standard deviation over $10$ seeds. For graph classification, we likewise use stratified $1{:}1{:}8$ splits over graphs and report mean and standard deviation over $10$ seeds. Node inputs are unified to 128 dimensions: COLLAB and IMDB-BINARY use degree-based one-hot features, while DD, ENZYMES, and PROTEINS use one-hot categorical node labels. Linear probing freezes the pretrained encoder; graph embeddings are obtained by mean pooling, $h_G=|V_G|^{-1}\sum_{i\in V_G}z_i$. For 1-shot node and graph classification, we sample one labeled example per class, keep the encoder frozen, and average results over $100$ independently sampled support sets.

\paragraph{Running environment.}
All experiments are conducted on a Linux server with 125 GiB RAM and 8 NVIDIA GeForce RTX 5090 GPUs, each with 32 GB memory. We use Python 3.9.23, PyTorch 2.8.0 with CUDA 12.8, and DGL 1.1.3.

\paragraph{Pre-training details.}
The model is pre-trained for 25 epochs. The encoder uses input, hidden, and output dimensions $128/1024/768$, three propagation layers, dropout $0.12$, heat times $[0.5,1.0,1.5]$, gate temperature $1.0$, and path-stream mix $1.0$. The spectral frontend applies $M_f=6$ Chebyshev filters of order $K_c=3$ to the normalized operator $S=D^{-1/2}AD^{-1/2}$ without self-loops. Each graph uses \(r_G=\min(32,n_G,d_{\mathrm{raw}})\) node-side truncated-SVD components per filter, zero-padded to 32 if needed, following the node-side SVD convention defined in the Method. With 6 filters, this gives a fixed 192-dimensional representation projected to 128 dimensions. Thus all graphs still produce the same \(M_f r\)-dimensional spectral representation before the final projection. The prototype module uses $p=96$ prototypes, diffusion time $t_{\mathrm{gw}}=0.58$, and GW entropy regularization $\epsilon_{\mathrm{gw}}=0.005$. The encoder is trained using the self-supervised objective described in Sec.~\ref{sec:self_supervised_pretraining}, with loss weights $1.5/0.3$ for node-level contrastive learning and spectral feature reconstruction, respectively. We optimize the model with Adam using learning rate $1.2\times10^{-4}$ and weight decay $10^{-5}$. Node contrast uses two edge-perturbed views and negative node pairs, masked reconstruction hides $25\%$ of nodes and reconstructs $X_{\mathrm{spec}}$.

\paragraph{\textbf{Spectral energy visualization.}}
All spectral visualizations are computed from graph-signal energy rather than classifier outputs. For graphs where eigendecomposition is feasible, we compute the normalized Laplacian \(L=I-D^{-1/2}AD^{-1/2}=U\Lambda U^\top\). Given a row-normalized representation \(X\), its normalized Fourier energy on eigenvector \(u_i\) is
\begin{equation}
E_i(X)=
\frac{\|u_i^\top X\|_2^2}
{\sum_j\|u_j^\top X\|_2^2}.
\label{eq:app_fourier_energy}
\end{equation}
We sort eigenvalues increasingly and sum the first, middle, and last thirds of the energies as low-, mid-, and high-frequency ratios. The ``spectral distribution after parsing'' figure uses \(X=X_{\mathrm{spec}}\), showing how Chebyshev filtering and SVD compression redistribute graph-signal energy. For large graphs, we use a residual-shell proxy: with \(P=D_{\tilde A}^{-1/2}\tilde A D_{\tilde A}^{-1/2}\) and \(\tilde A=A+I\), we set \(R_k=P^kX\), compute \(s_k=\|R_k-R_{k+1}\|_F^2\) for \(k=0,\ldots,7\) and \(s_8=\|R_8\|_F^2\). After normalization by \(\sum_{k=0}^{8}s_k\), we use \(s_0+s_1\), \(\sum_{k=2}^{7}s_k\), and \(s_8\) as high-, mid-, and low-frequency proxies.

\paragraph{\textbf{Prototype assignment visualization.}}
We visualize class-prototype associations to examine how different classes use the learned global prototype space. For each target dataset, we first compute a soft node-to-prototype transport matrix \(T\in\mathbb{R}^{N\times P}\) using the same KMeans-based transport procedure as in adaptation, where \(N\) is the number of nodes and \(P\) is the number of global prototypes. Each entry \(T_{i,p}\) measures the assignment strength between node \(i\) and prototype \(p\). We then row-normalize \(T\) to obtain per-node prototype assignment probabilities \(A\), so that each node has a probability distribution over prototypes. For each class \(c\), we average the assignment probabilities over all nodes belonging to that class:
\begin{equation}
M_{c,p}
=
\frac{1}{|\mathcal{V}_c|}
\sum_{i\in\mathcal{V}_c} A_{i,p},
\qquad
\mathcal{V}_c=\{i:y_i=c\}.
\label{eq:app_class_proto_matrix}
\end{equation}
The resulting matrix \(M\in\mathbb{R}^{C\times P}\) is visualized as a heatmap, where rows denote classes and columns denote global prototypes. Brighter entries indicate stronger association between a class and a prototype. 

\begin{figure}[t]
  \centering
  \captionsetup{font=small,skip=2pt}
  \captionsetup[subfigure]{font=footnotesize,skip=1pt}

  \begin{subfigure}[t]{0.48\columnwidth}
    \centering
    \includegraphics[width=\linewidth]{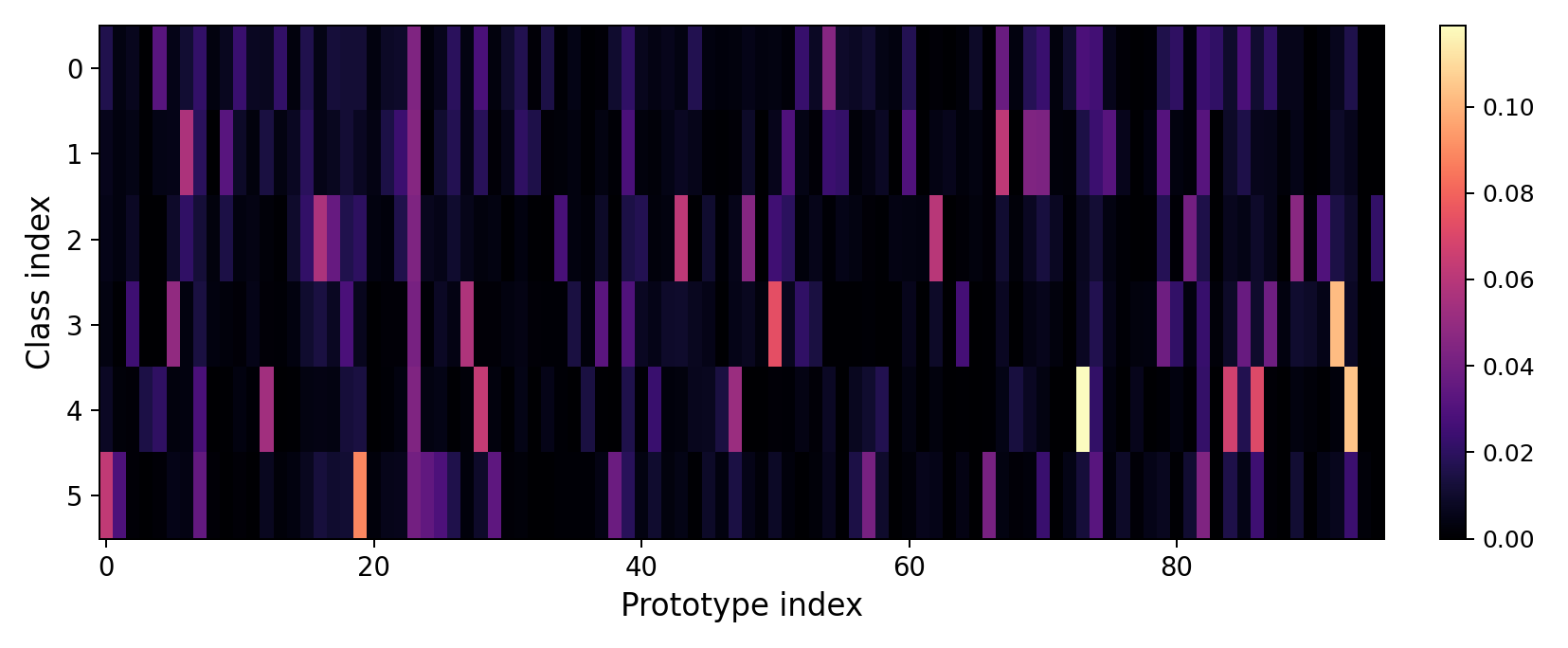}
    \caption{Citeseer}
    \label{fig:Citeseer_class_proto}
  \end{subfigure}
  \hfill
  \begin{subfigure}[t]{0.48\columnwidth}
    \centering
    \includegraphics[width=\linewidth]{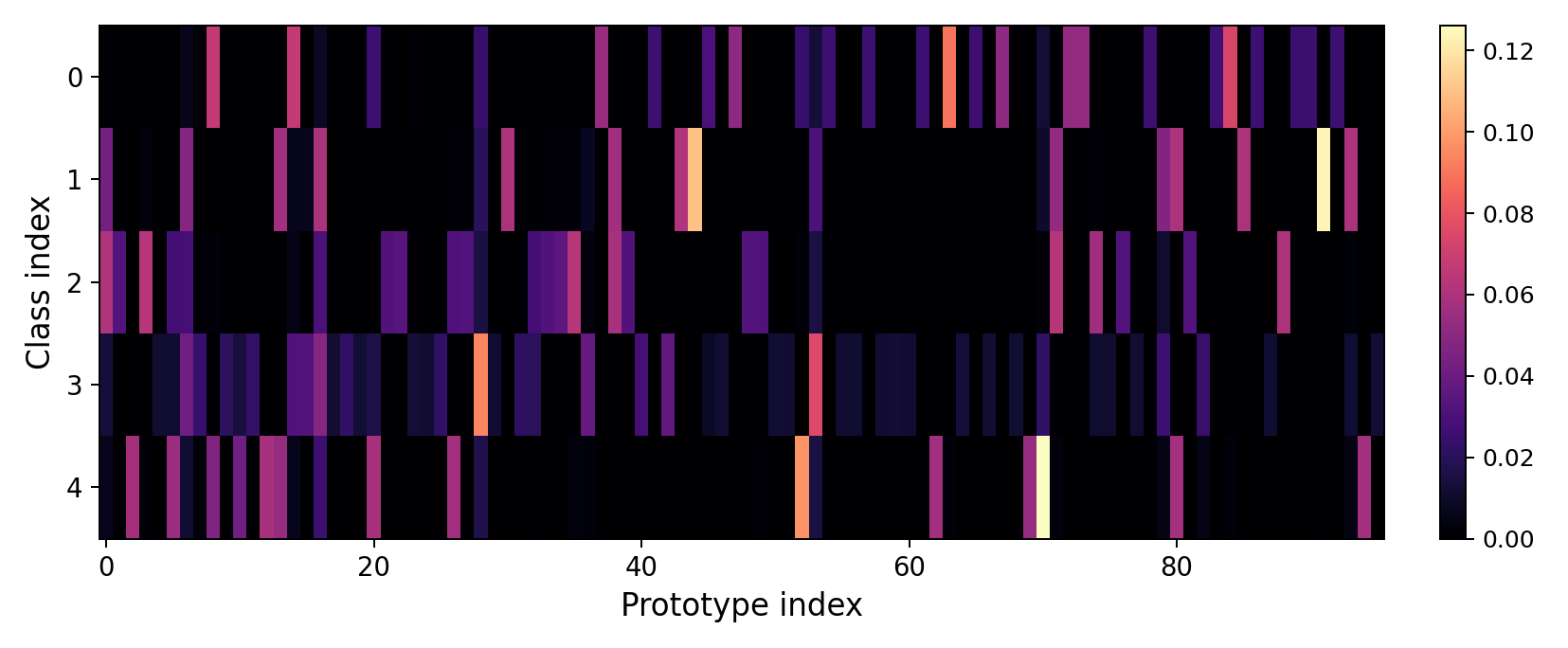}
    \caption{Cornell}
    \label{fig:Cornell_class_proto}
  \end{subfigure}

  \caption{Prototype assignment on Citeseer and Cornell.}
  \label{fig:class_proto}
\end{figure}

Figure~\ref{fig:class_proto} shows that different graphs and classes emphasize different subsets of the shared prototype space. We use this only as a qualitative view of prototype usage.

\paragraph{\textbf{Learned filter-response visualization.}}
The learned filter-response figure is computed directly from the trained Chebyshev coefficients, rather than from downstream predictions. Since the spectral frontend filters the normalized propagation operator \(S=D^{-1/2}AD^{-1/2}\), and \(S=I-L\) for the normalized Laplacian \(L\), a Laplacian frequency \(\lambda\in[0,2]\) corresponds to the propagation-domain value \(1-\lambda\). Therefore, when the same learned filter is visualized on the Laplacian-frequency axis, we write
\begin{equation}
\widehat g_\alpha(\lambda)
=
\sum_{k=0}^{K_c}\theta_{\alpha,k}T_k(1-\lambda),
\label{eq:app_filter_response}
\end{equation}
which is exactly the Method response \(g_\alpha(\rho)\) evaluated at \(\rho=1-\lambda\). We evaluate \(\widehat g_\alpha(\lambda)\) on a dense grid over \([0,2]\) and plot the normalized power \(|\widehat g_\alpha(\lambda)|^2\). To summarize each filter's dominant frequency preference, we also compute its power-weighted center. This visualization is used to verify whether the learned filters collapse to similar low-pass responses or instead cover complementary spectral regions.

\section{Datasets}

\begin{table}[H]
\centering
\caption{Dataset statistics. For graph classification datasets, Nodes and Edges are averages per graph.}

\label{tab:dataset_stats}
\small
\setlength{\tabcolsep}{4pt}
\renewcommand{\arraystretch}{1.08}
\begin{tabular}{llrrrrr}
\toprule
Dataset & Family & Graphs & Nodes & Edges & Node input & Classes \\
\midrule
\multicolumn{7}{l}{\textbf{Node Classification Datasets}} \\
Cora & Citation & 1 & 2,708 & 10,556 & 1,433 & 7 \\
Citeseer & Citation & 1 & 3,327 & 9,104 & 3,703 & 6 \\
Pubmed & Citation & 1 & 19,717 & 88,648 & 500 & 3 \\
Photo & Amazon co-purchase & 1 & 7,650 & 238,162 & 745 & 8 \\
Computers & Amazon co-purchase & 1 & 13,752 & 491,722 & 767 & 10 \\
CS & Coauthor & 1 & 18,333 & 163,788 & 6,805 & 15 \\
Physics & Coauthor & 1 & 34,493 & 495,924 & 8,415 & 5 \\
Cornell & WebKB & 1 & 183 & 298 & 1,703 & 5 \\
Wisconsin & WebKB & 1 & 251 & 515 & 1,703 & 5 \\
Chameleon & WikipediaNetwork & 1 & 2,277 & 36,101 & 2,325 & 5 \\
ogbn-arxiv & OGB citation & 1 & 169,343 & 1,166,243 & 128 & 40 \\
ogbn-products & OGB co-purchase & 1 & 2,449,029 & 61,859,140 & 100 & 47 \\
\midrule
\multicolumn{7}{l}{\textbf{Graph Classification Datasets}} \\
COLLAB & Social & 5,000 & 74.49 & 2,457.78 & Degree & 3 \\
DD & Bioinformatics & 1,178 & 284.32 & 715.66 & Label & 2 \\
ENZYMES & Bioinformatics & 600 & 32.63 & 62.14 & Label & 6 \\
IMDB-B & Social & 1,000 & 19.77 & 96.53 & Degree & 2 \\
PROTEINS & Bioinformatics & 1,113 & 39.06 & 72.82 & Label & 2 \\
\bottomrule
\end{tabular}
\end{table}

For graph classification, ``Degree'' denotes degree-based one-hot node inputs used for the featureless COLLAB and IMDB-B datasets, while ``Label'' denotes one-hot encoded categorical node labels used as node features for DD, ENZYMES, and PROTEINS. All resulting node inputs are unified to 128 dimensions before being fed into SPR.

The node-classification benchmarks span citation, co-purchase, co-authorship, WebKB/Wikipedia, and large-scale OGB graphs; graph classification further evaluates transfer to whole-graph prediction. Dataset statistics are reported in Table~\ref{tab:dataset_stats}.
\section{Complexity Analysis}
\label{app:complexity}
Let \(n\) and \(e\) be the number of nodes and edges of an input graph, \(d_0\) the raw feature dimension, \(d\) the unified spectral dimension, \(h\) the hidden dimension, \(d_z\) the output dimension, \(M_f\) the number of spectral filters, \(K_c\) the Chebyshev order, \(r\) the SVD rank, \(p\) the number of prototypes, and \(L\) the number of propagation layers.

\paragraph{Spectral parsing.}
The Chebyshev frontend applies \(M_f\) filters of order \(K_c\). Since each Chebyshev recursion requires sparse multiplication with the normalized propagation operator \(S\), the cost of computing all filter responses is \(O(M_fK_ced_0)\). For each filtered response, rank-\(r\) truncated SVD costs approximately \(O(nd_0r)\), giving \(O(M_fnd_0r)\) over all filters. The final projection from the concatenated \(M_fr\)-dimensional representation to \(d\) dimensions costs \(O(nM_frd)\). Therefore, the overall spectral parsing cost is
\begin{equation}
O\!\left(M_fK_ced_0 + M_fnd_0r + nM_frd\right).
\label{eq:app_spectral_parsing_complexity}
\end{equation}
The memory cost is \(O(e+nd_0+nM_fr)\), plus learnable parameters.

\paragraph{Prototype construction.}
For each source graph \(G_m\), constructing a dense heat-kernel-distance matrix requires \(O(n_m^2)\) memory. Exact heat-kernel computation by eigendecomposition costs \(O(n_m^3)\), while the implementation uses an approximate normalized-adjacency surrogate for large graphs. In each GW update between a source graph and \(p\) prototypes, the linearized cost matrix costs \(O(n_m^2p+n_mp^2)\), and Sinkhorn normalization costs \(O(n_mp)\) per iteration. With \(B\) barycenter outer iterations, \(G\) GW iterations, and \(S\) Sinkhorn iterations, the dominant prototype construction cost is
\begin{equation}
O\!\left(
B\sum_m G(n_m^2p+n_mp^2)
+
B\sum_m GS\,n_mp
\right).
\label{eq:app_prototype_construction_complexity}
\end{equation}
This cost is paid during pre-training and is amortized across downstream tasks.

\paragraph{Gated propagation.}
At each propagation layer, the adjacency path costs \(O(eh)\). If the heat kernel is explicitly materialized, the heat path costs \(O(n^2h)\); for large graphs, the implementation instead uses sparse or approximate propagation. The prototype path is defined by \(P=TK_\star T^\top\). Explicitly forming \(P\) would cost \(O(np^2+n^2p)\), and applying it would cost \(O(n^2h)\). In practice, the operator is applied implicitly as \(T(K_\star(T^\top H))\), which costs \(O(nph+p^2h)\) per layer, or \(O(nkh+p^2h)\) when each node keeps only \(k\) nonzero prototype assignments. The node-wise gate and hidden transformations cost \(O(nh^2)\) when hidden-to-hidden projections dominate. Under sparse adjacency and sparse prototype transport, the total \(L\)-layer propagation cost is
\begin{equation}
O\!\left(
L(eh+nkh+p^2h+nh^2)
\right).
\label{eq:app_gated_propagation_complexity}
\end{equation}

\paragraph{Adaptation to unseen graphs.}
For a regular-sized target graph, GW-based assignment has the same graph-to-prototype complexity as above. For a large target graph, the scalable K-means-based assignment costs \(O(Inpd_t)\), where \(I\) is the number of iterations and \(d_t\) is the assignment feature dimension, followed by top-\(k\) sparsification. The downstream linear classifier costs \(O(nd_zc)\) per epoch for \(c\) classes. The large-graph approximation is used for scalability and is not covered by the GW transport guarantee above.

\begin{table}[t]
\centering
\caption{Runtime and GPU profiling results.}
\label{tab:runtime_gpu_profile}
\small
\setlength{\tabcolsep}{5pt}
\renewcommand{\arraystretch}{1.08}
\begin{tabular}{lccccc}
\toprule
Dataset & Total Time & Forward Time & Linear Time & Transport Time & Peak GPU \\
\midrule
ogbn-arxiv & 12.5s & 0.7s & 3.1s & 1.1s & 3037 MiB \\
ogbn-products & 53m 8.1s & 51m 51.2s & 1m 11.1s & 5.8s & 32099 MiB \\
Cora & 8.4s & 0.2s & 1.8s & 0.6s & 911 MiB \\
Physics & 1m 3.1s & 0.8s & 53.1s & 2.1s & 14307 MiB \\
\bottomrule
\end{tabular}
\end{table}

\paragraph{Large-graph handling strategy.}
For large-scale graphs, we avoid dense $N\times N$ pairwise matrices and exact full-graph GW alignment. For ogbn-arxiv and ogbn-products, we keep the learned prototype geometry fixed and replace only the target graph-to-prototype alignment with a K-means-based sparse assignment using 96 slots and top-$k$ sparsification, together with sparse adjacency. This is a scalable approximation to the exact GW assignment and is not covered by Proposition~\ref{prop:gw_lifting}. For ogbn-products, inference uses sampled subgraphs with batch size 1024 and fanouts $(10,10,5)$, followed by linear probing on frozen embeddings.

\end{document}